\documentclass[10pt]{article}
\usepackage[accepted]{tmlr}

\usepackage{amsmath,amsfonts,bm}

\newcommand{\figleft}{{\em (Left)}}
\newcommand{\figcenter}{{\em (Center)}}
\newcommand{\figright}{{\em (Right)}}

\def\eqref#1{equation~\ref{#1}}

\def\1{\bm{1}}

\DeclareMathAlphabet{\mathsfit}{\encodingdefault}{\sfdefault}{m}{sl}
\SetMathAlphabet{\mathsfit}{bold}{\encodingdefault}{\sfdefault}{bx}{n}

\usepackage{hyperref}
\usepackage{url}
\usepackage{booktabs}
\usepackage{multirow}
\usepackage{graphicx}
\usepackage{amsmath,amssymb}
\usepackage{combelow}

\newcommand{\sbn}{\ensuremath{\sigma}\textnormal{BN}}
\newcommand{\sln}{\ensuremath{\sigma}\textnormal{LN}}
\newcommand{\sn}{\ensuremath{\sigma}\textnormal{Norm}}
\newcommand{\snens}{\ensuremath{\sigma}\textnormal{N-Ens}}
\newcommand{\Ldiv}{\mathcal{L}_{\mathrm{div}}}
\newcommand{\softmaxt}{\mathrm{softmax}_{\tau}}

\newcommand{\sncos}{\ensuremath{\sigma}\textnormal{Norm-cos}}
\newcommand{\Umod}{U_{\mathrm{mod}}}
\newcommand{\thetaconv}{\theta_{\mathrm{shared}}}
\newcommand{\Ent}{H}
\newcommand{\MI}[1]{I[#1]}

\title{Controllable Diversity in Normalization-Based \\ Implicit Ensembles via Softmax-Temperature Modulation}

\author{\name Mihai \cb Suteu \email{m.suteu16@imperial.ac.uk}\\
\addr Data Science Institute, Imperial College London\\
\AND
\name Ovidiu \cb Serban \email{o.serban@imperial.ac.uk}\\
\addr Data Science Institute, Imperial College London\\
}

\def\month{10}
\def\year{2026}
\def\openreview{\url{https://openreview.net/forum?id=Jk5B1JFrPK}}

\begin{document}

\maketitle

\begin{abstract}
Deep ensembles provide the most reliable uncertainty estimates in deep learning, but their cost scales linearly with the number of members. Implicit ensembles mitigate this cost by sharing a single backbone across members. Member diversity is a primary determinant of their quality, yet existing implicit ensembles fix it at initialisation or build it into the architecture, without direct control during training. We introduce \snens{}, a normalisation-based implicit ensemble that treats each member as a task in a multi-task architecture and modulates the shared backbone through sigmoid-bounded scalers. Using a softmax regulariser on these scalers lets a single temperature parameter shape the equilibrium level of sharing between members and move the ensemble along the accuracy-calibration tradeoff. We further show that this parameter-space diversity tracks output diversity, so it can be read cheaply from the weights. Because only normalisation layers are replicated, the same mechanism can wrap convolutional and transformer backbones, and can convert a pretrained model into a calibrated ensemble through a short fine-tune. We frame the epistemic uncertainty that such an ensemble expresses as modulation uncertainty and explain why its calibration holds under input corruption and why its out-of-distribution detection is weaker. We evaluate \snens{} across ResNets and transformers on CIFAR-10/100, ImageNet, and SST-2. \snens{} is competitive with state-of-the-art methods, scales with ensemble size, maintains calibration under distribution shift, and lets member diversity be controlled during training. Code available at: \url{https://github.com/xapharius/TSsNorm}

\end{abstract}

\section{Introduction}
Uncertainty quantification is a prerequisite for deploying deep learning in risk-sensitive domains \citep{gawlikowski2022, he2025}. Deep ensembles \citep{lakshminarayanan2017} remain the strongest generic approach, as independently trained members converge to distinct modes of the loss landscape \citep{fort2019}, and their disagreement on novel inputs provides a calibrated epistemic signal. Their cost scales linearly with the number of members, which becomes prohibitive as model size grows. Implicit ensembles \citep[e.g.][]{wen2020batchensemble, laurent2022packed} address this by sharing a single backbone across members and introducing per-member modulation or partitioning. They can achieve performance comparable to deep ensembles on standard benchmarks at a fraction of the parameter cost.

Within an ensemble, a primary determinant of quality is the diversity of its members. \citet{wood2023bvd} show that for any Bregman loss, the expected ensemble loss decomposes into member bias, member variance, and diversity. Diversity is a tradeoff rather than a quantity to maximise, and \citet{abe2024pathologies} confirms that at high capacity, forcing members apart harms performance. The right level of diversity is therefore something to choose, yet implicit ensembles offer no direct control over it. In adapter-based methods \citep{wen2020batchensemble, turkoglu2022film} diversity is set at initialisation and drifts freely thereafter; in partitioning-based methods \citep{laurent2022packed, durasov2021masksembles} it is built in structurally, at the expense of individual member capacity. Neither lets the level of member disagreement be set during training.

We reinterpret implicit ensembles through the multi-task learning paradigm, which concerns capacity allocation when resources are shared. We adopt TS\sbn{} \citep{tssbn}, an MTL architecture that confines modulation to the normalisation layers and leaves the full feature extractor available to every member. The modulation primitive is \sbn{} \citep{sigmabn}, which only uses sigmoid-bounded scalers and can be interpreted as feature importances. We use it in a general \sn{} form that applies to any normalisation layer. These feature level importances make member diversity measurable as the cosine similarity between member importance vectors, which a regulariser can act on. Members are then modulated independently and can predict alike but for different reasons \citep{rame2021dice}.

To control diversity, we introduce a softmax regulariser applied per feature across members' \sn{} importance vectors. A single temperature $\tau$ sets the equilibrium to which the importances settle, controlling how much each feature is shared across members. Unlike a conventional regularisation strength, which tunes how hard the network is pushed toward one fixed target, $\tau$ selects among a family of equilibria and holds diversity at the chosen level throughout training. Diversity is thus controlled during training, rather than fixed at initialisation or by construction as in existing methods.

The resulting method matches strong implicit and deep-ensemble baselines across vision and language benchmarks. It improves performance as ensemble size grows, whereas partitioning methods degrade as individual member capacity decreases. By sweeping $\tau$, the ensemble moves along an accuracy-calibration tradeoff in which the parameter-space diversity metric tracks output-space diversity (\S\ref{sec:tradeoff}). Because it can act on any normalisation layer, our method works on both convolutional and transformer backbones. In addition it can turn an existing pretrained model into a calibrated ensemble with a short fine-tune as an alternative to retraining members from scratch.
Our contributions:
\begin{itemize}
    \item \textbf{Controllable diversity.} We introduce a softmax regulariser on \sn{} importance vectors. A single temperature $\tau$ sets the level of member diversity and moves the ensemble along the accuracy-calibration tradeoff. To our knowledge, this is the first mechanism to control member diversity during backbone training in the implicit ensemble family, rather than fixing it at construction or initialisation. We further show this parameter-space diversity tracks the output diversity that governs ensemble quality (\S\ref{sec:tradeoff}).

    \item \textbf{Modulation uncertainty.} We split the ensemble's epistemic term and identify modulation uncertainty as the part an ensemble with a shared backbone expresses by reweighting features. This predicts that calibration holds under input corruption while out-of-distribution detection is weaker (\S\ref{sec:modunc}).

    \item \textbf{Competitiveness across architectures and regimes.} Because only the normalisation layers are swapped, \sn{} can wrap any standard backbone and can convert a single pretrained model into a calibrated ensemble through a short fine-tune. Across ResNets and transformers on CIFAR-10/100, ImageNet, and SST-2, \snens{} is competitive with state-of-the-art methods, improves with ensemble size where partitioning methods collapse, and maintains calibration under distribution shift.

\end{itemize}

\section{Related Work}\label{sec:relwork}
\textbf{Uncertainty quantification} in deep learning spans three families: Bayesian approaches \citep{blundell2015, ritter2018, wellingteh2011}, single model deterministic methods \citep{liu2020sngp, sensoy2018, osband2022}, and deep ensembles \citep{lakshminarayanan2017}. Deep ensembles remain the strongest under distribution shift, but scale linearly in the number of members. Post-hoc calibration, such as temperature scaling \citep{guo2017}, is orthogonal and cheaper but improves neither accuracy nor epistemic coverage and relies on a single scale whose calibration degrades under distribution shift \citep{ovadia2019}.

\textbf{Implicit ensembles} reduce this cost by sharing a single backbone and adding per-member modulation or partitioning, producing ensemble-like behaviour in a single training run at a fraction of the parameter count. We organise the family by whether its members arise from architectural partitioning or from modulation of a shared backbone. Partitioning methods carve the architecture into sub-models that are nearly independent: Packed Ensembles \citep{laurent2022packed} hard-partition the channel space, Masksembles \citep{durasov2021masksembles} mask features, and MIMO \citep{havasi2021mimo} split the input and output projections. Because the partition is built into the architecture, these methods require a bespoke redesign for each model family and cannot be applied to a pretrained backbone. Modulation methods instead keep the full backbone and inject per-member adapters: BatchEnsemble \citep{wen2020batchensemble} uses rank-1 perturbations and LoRA-Ensemble \citep{muhlematter2025lora} low-rank attention adapters. However, BatchEnsemble's functional diversity is near zero at moderate scale \citep{zamyatin2026}. FiLM-Ensemble \citep{turkoglu2022film} is our closest neighbour, modulating a shared backbone with per-member BatchNorm parameters. Still, its diversity is fixed by the variance of its initialisation, which confines it to training from scratch. Across both branches, diversity is a byproduct of the design. Its level is set at initialisation or built into the architecture, and cannot be chosen during backbone training. A detailed comparison to related models can be found in ~\ref{sec:app-baselines}.

\textbf{Learned diversity} mechanisms have been developed at three levels of the network: output-space regularisers on member predictions \citep{liuyao1999, pang2019, mehrtens2022}, feature-space objectives on member representations \citep{rame2021dice, sinha2020}, and parameter-space repulsion on member weights \citep{dangelo2021, steger2024}. \citet{abe2024pathologies} show that enforcing output diversity can harm performance at high capacity, though their analysis excludes the other two levels. These mechanisms were developed for explicit or branched ensembles, whose members carry separate weights to diversify. The nearest work in the implicit ensemble family is \citet{kirsch2025}, which extracts diverse sub-models from a frozen backbone under a parameter-space diversity objective. With the backbone fixed and no control over the level of diversity, it diversifies once rather than maintaining a chosen level during backbone training. Separately, the aleatoric-epistemic decomposition of predictive uncertainty is increasingly contested. The components are hard to measure, and the estimators meant to isolate them turn out to be strongly conflated \citep{mucsanyi2024, bickfordsmith2025}. We broadly agree, and rather than claiming to recover a ``true'' epistemic quantity, we focus on defining what such a modulation mechanism can express in practice (\S\ref{sec:modunc}).

\textbf{Softmax temperature} is applied through our regulariser over static weight parameters, a use distinct from its normal role in deep learning. Its established uses act on other objects: it controls the entropy of an output distribution, as in distillation, entropy regularised policies, and confidence penalties \citep{hinton2015, haarnoja2018, pereira2017}; it gates activations recomputed each forward pass, as in attention and mixtures of experts \citep{vaswani2017, shazeer2017, fedus2022}; and it relaxes discrete architectural choices that resolve by argmax at the end of training \citep{liu2019darts, jang2017, maddison2017, sun2020adashare}. In every case, the controlled distribution is defined over outputs, activations, or latent choices that collapse to a discrete selection. To our knowledge, softmax temperature has not otherwise been used as a regulariser on the weights themselves.

\textbf{Normalisation layers} have been used as lightweight sites for uncertainty, such as sampling BatchNorm statistics for Bayesian-style uncertainty \citep{atanov2019}, and post-hoc Bayesian adaptation through them \citep{franchi2024abnn}. FiLM-Ensemble \citep{turkoglu2022film} is the most direct precedent, modulating a shared backbone with per-member BatchNorm parameters, drawn from the multi-task treatment of normalisation. We build on \sbn{} \citep{sigmabn}, a sigmoid-based primitive trained with differential learning rates and used for capacity allocation in MTL \citep{tssbn}. Its scales are amenable to regularisation, which we use to shape diversity across members independently of initialisation.

\textbf{Pretrained transformers} being converted to implicit ensembles is a recent direction the field has taken. LoRA-Ensemble \citep{muhlematter2025lora} adds low-rank attention adapters per member with diversity from random initialisation, and Hydra Ensembles \citep{gabetni2026hydra} duplicates and fuses heads under a pruning protocol. Both are specific to transformers, and neither controls diversity. LoRA-Ensemble keeps the backbone frozen and diversifies through adapter initialisation, while Hydra fine-tunes its duplicated heads. Instead, \sn{} replaces the normalisation layers of either architecture family, with diversity shaped through training.

\section{Method}\label{sec:method}
\subsection{Architecture}\label{sec:arch}
We adopt the multi-normalisation construction of TS\sbn{} \citep{tssbn}, which replaces each shared normalisation layer with a separate layer per task. We restate it for general normalisation, since the same construction was shown to work over both BatchNorm and LayerNorm:
\begin{equation}
\sigma\text{Norm}_t(x) = \sigma(\gamma_t)\odot\text{norm}_t(x), \qquad \sigma(\gamma_t)=\frac{1}{1+e^{-\gamma_t}},
\end{equation}
where $\text{norm}_t$ is task $t$'s own normalisation step and $\gamma_t$ its scale logits. The primitive \sn{} bounds the scaler to $(0,1)$ and admits a differential learning rate on $\gamma$ for which we adopt the TS\sbn{} default. Since the focus is on the affine transformation, it can wrap existing normalisation layers: \sbn{} over BatchNorm, \sln{} over LayerNorm. We treat each task as an ensemble member: $M$ members share every convolution or attention weight, differ only in their \sn{} scales, and each carries its own initialised head. Each member requires a different forward pass. However, this can be vectorised into a single batched pass, as introduced by BatchEnsemble \citep{wen2020batchensemble}.

Treating each member as a task imports the MTL toolkit of \citet{tssbn}. The sigmoid bounds each scale to a readable per-feature importance, so member disagreement can be measured as the cosine similarity between importance vectors. This is a label-free diversity metric computed from the parameters alone. Because diversity comes from swapping normalisation layers, a single primitive wraps any backbone without redesigning the architecture, as partitioning methods or CNN-specific adapters require (\S\ref{sec:relwork}).

While the MTL framing provides the tools, the training dynamics and tension are inverted compared to the ensemble setting. In MTL tasks, distinct objectives have conflicting gradients, and the machinery exists to reduce interference \citep{yu2020pcgrad}. In our construction, every member optimises the same objective on the same data, so the loss creates no pressure to differentiate; diversity must be supplied externally. Seeding it through initialisation alone, as prior modulation ensembles do, fixes it at the start with no control over it thereafter as members drift. A mechanism that enforces diversity during training can control the level regardless of initialisation, including pretrained networks (Appendix~\ref{sec:app-pretrained}). The MTL bias toward generic shared features \citep{caruana1997} still applies to the backbone but, on its own, does not yield a controlled level of member diversity, which \S\ref{sec:mechanism} supplies.

\subsection{Controllable diversity via softmax-temperature regularisation}\label{sec:mechanism}
\begin{figure}[t]
\begin{center}
\includegraphics[width=\linewidth]{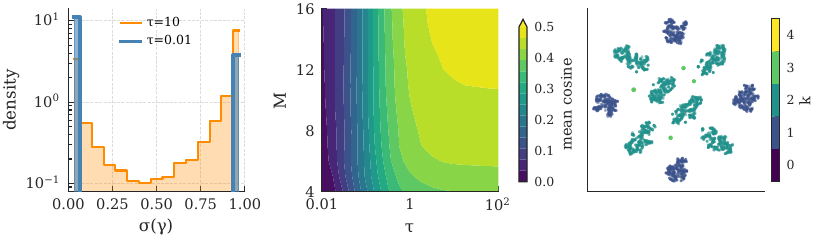}
\end{center}
\caption{\figleft{} per-feature importance distributions for ResNet18, showing concentration toward the sigmoid extremes as temperature is decreased. \figcenter{} simulated mean \sn{}-cosine over the different member sizes and temperature values. \figright{} clustering of features per their member importance vectors, showing the $M$-choose-$k$ allocation.}
\label{fig:composite}
\end{figure}

Since members share a backbone and optimise the same loss on the same data, nothing in the objective pushes them apart, so the regulariser must supply that pressure. The goal is to assign each feature to only some members, rather than making it fully shared and thus generic. For a given feature, we look at how strongly each member uses it and push that usage to concentrate on a few members. A single temperature $\tau$ sets how concentrated the usage becomes. At low $\tau$ a feature is owned by a single member, the hard partition that partitioning methods build in by construction, and at high $\tau$ it is shared across all members. Between these, a feature is owned by an effective $k$ of the $M$ members, a soft $M$-choose-$k$ allocation, and $\tau$ sets how many members own each feature. Figure~\ref{fig:composite}~\figright{} shows this, with features clustering into groups owned by $k$ members (those with $\sigma(\gamma) > 0.5$). The implicit ensemble family has so far occupied only the two endpoints, partitioning at $k=1$ and shared modulation at $k=M$. $\tau$ lets us select the points between them. Because it acts through a regulariser, that point is set and held during training for any backbone, rather than fixed at construction or initialisation.

Concretely, for each regularised layer $\ell$ and feature $i$, we collect the $M$ members' importances $\sigma(\gamma_{1,\ell,i}),\dots,\sigma(\gamma_{M,\ell,i})$ and take a softmax across members at temperature $\tau$,
\begin{equation}
\Ldiv = \lambda \sum_{\ell\in\mathcal{R}}\ \sum_{i}\ \sum_{m=1}^{M} \log\big[\softmaxt\big(\sigma(\gamma_{\cdot,\ell,i})\big)\big]_m,
\end{equation}
where $\mathcal{R}$ is the set of regularised layers, and the softmax is taken over the member index for one feature. For the vector $z$ of the $M$ members' importances $[\softmaxt(z)]_m = e^{z_m/\tau}\big/\sum_{m'=1}^{M} e^{z_{m'}/\tau}$. Minimising this sharpens each feature's across-member distribution, so the more a single member owns a feature, the lower the term, and members are driven to differentiate in which features they rely on. The temperature $\tau$ sets how sharp that target distribution is.

This is where $\tau$ and the strength $\lambda$ play distinct roles. $\tau$ sets the attractor the importances are pulled toward, while $\lambda$ sets how strongly they are pulled there against the task loss. Unlike a monotone penalty, where the strength alone fixes the endpoint, our softmax penalty defines a $\tau$-parametrised target that selects which equilibrium the system moves toward. Trained from scratch, the endpoint is then governed by $\tau$ and is empirically insensitive to $\lambda$, which we keep fixed across experiments. We ablate the impact of $\lambda$ in \S\ref{sec:lambda}.

The two temperature limits bound the behaviour and are derived in Appendix~\ref{sec:app-limits}. As $\tau\to 0$, the softmax target becomes one-hot, and each feature is driven to a single owner as a maximally diverse state. As $\tau\to\infty$, it becomes uniform, which carries no gradient toward any particular allocation, and the regulariser stops pushing. Figure~\ref{fig:composite}~\figleft{} shows the effect for a ResNet18 during training. Under sustained pressure at low temperature, the importances move toward the extremes of the sigmoid.

Between the limits, $\tau$ moves the diversity level monotonically, as illustrated in Figure~\ref{fig:composite}~\figcenter{}. We isolate the effect of the regulariser in a synthetic setting across a grid of ensemble size $M$ and temperature $\tau$. Using importance matrices initialised from $\mathcal{N}(0,1)$, the same initialisation used in training, we plot the mean pairwise cosine between members once the member assignment stabilises. The surface is smooth and monotonic in $\tau$ and varies only weakly with $M$, so $\tau$ moves the diversity level largely independently of ensemble size. Full simulation details are in Appendix~\ref{sec:app-limits}.

\section{Analysis}\label{sec:analysis}
The previous section introduces a simple mechanism that controls the parameter-space diversity among members via regularisation. For the mechanism to be useful, the parameter-space diversity it induces must translate into output diversity, the quantity that enters the performance tradeoff (\S\ref{sec:tradeoff}). Similarly, because the backbone is shared among members, we need to understand how this affects the overall uncertainty the mechanism can express and to characterise its strengths and weaknesses (\S\ref{sec:modunc}).

\subsection{The diversity tradeoff}\label{sec:tradeoff}
Diversity is the reason an ensemble outperforms its individual members. Disagreement on hard or shifted inputs drives calibration and supplies the epistemic uncertainty an ensemble is meant to provide. But naively adding more diversity does not necessarily translate into better performance. For any Bregman loss $\mathcal{L}$, including cross-entropy, \citet{wood2023bvd} decompose the expected ensemble loss exactly into
\begin{equation}
\mathbb{E}_D[\mathcal{L}(y, q^*)] = \underbrace{\tfrac{1}{M}\sum_{i} \mathcal{L}(y, \bar{q}_i)}_{\text{avg.\ bias}} + \underbrace{\tfrac{1}{M}\sum_{i} \mathbb{E}_D[\mathcal{L}(\bar{q}_i, q_i)]}_{\text{avg.\ variance}} - \underbrace{\tfrac{1}{M}\sum_{i} \mathbb{E}_D[\mathcal{L}(q^*, q_i)]}_{\text{diversity}},
\end{equation}
with $q_i$ member $i$'s prediction, $\bar{q}_i$ its Bregman centroid over $D$, and $q^*$ the corresponding centroid combiner across members. Diversity enters negatively, but the three terms are coupled, so pushing members apart raises diversity but also perturbs bias and variance. The best level of diversity is therefore a tradeoff. We show empirically that this tradeoff is visible for the parameter-space diversity induced by the temperature $\tau$.

\begin{figure}[t]
\begin{center}
\includegraphics[width=0.48\linewidth]{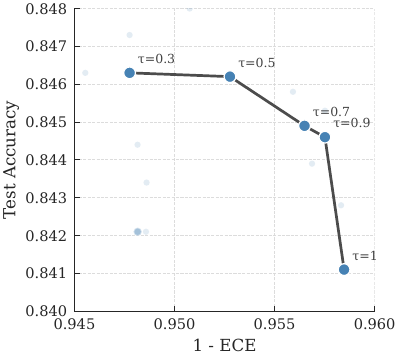}\hfill
\includegraphics[width=0.48\linewidth]{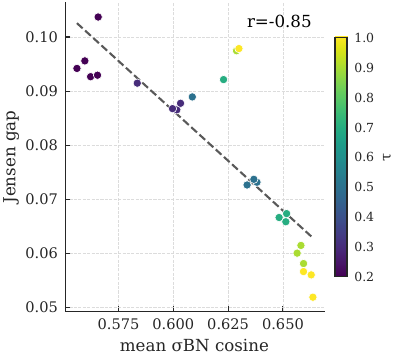}
\end{center}
\caption{\figleft{} accuracy-calibration tradeoff over a $\tau$ sweep on WideResNet-28-10 / CIFAR-100, 3 seeds, with $\tau$ moving the ensemble monotonically along it. \figright{} \sn{}-cosine against Jensen gap across the same sweep.}
\label{fig:pareto}
\end{figure}

Sweeping $\tau$ yields monotonic movement along an accuracy-calibration tradeoff. Figure~\ref{fig:pareto}~\figleft{} shows this on WideResNet-28-10 over 3 seeds on CIFAR-100. Training can then be guided toward a performance level that matches an application's risk sensitivity.

The decomposition itself rewards \emph{output} diversity, where members predict differently when it helps, or predict alike for different reasons \citep{rame2021dice}. Our regulariser, however, acts on \emph{parameters}, the \sn{} importances. We need the two to coincide, and the bounded sigmoid primitive lets us check whether they do. We define the parameter-space diversity as the pairwise cosine between member importance vectors, which is
\begin{equation}
\sncos = \operatorname*{mean}_{\ell,\ i<j}\, \cos\!\big(\sigma(\boldsymbol{\gamma}^{\ell}_i),\, \sigma(\boldsymbol{\gamma}^{\ell}_j)\big),
\end{equation}
computed from the weights alone, with no forward pass or labels. Its output-space counterpart is the Jensen gap \citep{abe2022}, the diversity term under the arithmetic-mean combiner \citep{wood2023bvd},
\begin{equation}
\mathcal{J} = \tfrac{1}{M}\sum_{i} \mathcal{L}(y, q_i) - \mathcal{L}(y, \bar{q}),
\end{equation}
the average member loss minus the loss of the averaged prediction is non-negative and exactly zero when members agree. Across $\tau$ the two track closely (Figure~\ref{fig:pareto}~\figright{}, $|\rho|\approx0.85$; $|\rho|\ge0.83$ on ResNet-18, ResNet-50, and WideResNet). The cheap parameter-space quantity is therefore a faithful proxy for the output diversity the framework rewards and can be monitored without a forward pass. The relationship holds for other output diversity measures, which we show in Appendix~\ref{sec:app-fourmetric}.

\subsection{Modulation uncertainty}\label{sec:modunc}
All members share the backbone, and therefore the same feature extractors, meaning they differ only in how they reweight the features those extractors produce. This limits the uncertainty such an ensemble can express. Two members may disagree about how much to rely on a given feature, but not about which features to learn. The disagreement expressed by the ensemble is therefore only a subset of that expressed by fully independent members, omitting the component that arises from different feature extraction. This subsection formalises that limitation and explains its implications for calibration and out-of-distribution detection.

The diversity an ensemble exposes is a form of epistemic uncertainty. Predictive uncertainty at an input $x$ is the entropy $\Ent[Y\mid x]$ of the ensemble's mean prediction; it splits into an aleatoric part, the mean entropy of the individual members, and an epistemic part, the mutual information $\MI{Y;M\mid x}$ between the prediction and the member index $M$ \citep{smithgal2018}. Refining this decomposition by conditioning on part of the model is an established device \citep{depeweg2018}, and confining epistemic uncertainty to a small parameter subset over a deterministic backbone has precedents \citep{dusenberry2020, trinh2022, daxberger2021}. Our construction is similar: member $m$ predicts $f(y\mid x, \thetaconv, \phi_m)$, sharing every backbone weight and varying only the \sn{} importances $\phi_m$. For the idealised ensemble in which both parts vary, the chain rule of mutual information splits the epistemic term exactly and without independence assumptions,
\begin{equation}\label{eq:chain-main}
\MI{Y;(\theta,\phi)\mid x} \;=\; \underbrace{\MI{Y;\theta\mid x}}_{\text{backbone}} \;+\; \underbrace{\MI{Y;\phi\mid x,\theta}}_{\text{modulation}}.
\end{equation}
Our ensemble realises the second component at the single draw $\theta=\thetaconv^\star$, which we call modulation uncertainty, $\Umod(x) = \MI{Y;\phi\mid x,\ \theta=\thetaconv^\star}$, and structurally forgoes the first. Because $\Umod$ is one realisation of the conditional term rather than the term itself, it cannot be recombined with any measured quantity to recover the full epistemic term (Appendix~\ref{sec:app-modunc}).

The decomposition is structural and does not say how large each component is at a given input. Those follow from an empirical premise: independently trained backbones agree on in-distribution inputs and diverge as inputs move out of distribution \citep{fort2019, ovadia2019}. Equation~\ref{eq:chain-main} then makes two predictions ahead of \S\ref{sec:results}. A corrupted input is one the backbone still represents, so the modulation component carries calibrated disagreement and calibration should degrade slowly under shift. An out-of-distribution input is one the backbone cannot represent, so the disagreement that would detect it lies in the forgone backbone component, and detection should be weak. \S\ref{sec:results} confirms both on the pretrained vision backbones, where the two sides appear in a single model. In the from-scratch regime, where members shape the shared representation during training, calibration under shift holds while detection instead improves with capacity.

\section{Experiments}\label{sec:experiments}
\subsection{Setup}\label{sec:setup}
All experiments are run within TorchUncertainty \citep{franchi_tu}. For the convolutional networks, we adopt the Packed Ensembles protocol \citep{laurent2022packed} and our results append directly to the published Packed Ensembles tables. For the transformers, we follow the Hydra Ensembles protocol \citep{gabetni2026hydra}, which is implemented on an unreleased branch of TorchUncertainty. On SST-2, where Hydra uses the public BERT-base backbone, we keep it as a baseline and append our row to its table. Its ViT results, however, rely on a bespoke ImageNet-21k backbone that is not publicly released and therefore cannot be reproduced. We thus rerun ViT-B/16 benchmarks on public timm backbones using a shared recipe.

We evaluate on CIFAR-10/100 \citep{krizhevsky2009} with ResNet-18/50 \citep{he2016resnet} and WideResNet-28-10 \citep{zagoruyko2016wrn}, on SST-2 \citep{socher2013sst} with BERT-base \citep{devlin2019bert}, and on CIFAR-100 and ImageNet \citep{deng2009imagenet} with ViT-B/16 \citep{dosovitskiy2021vit}. We report predictive quality (accuracy, NLL as the primary proper score, Brier on the transformer tables), calibration (ECE, and adaptive calibration error, aECE, where available), out-of-distribution detection (AUROC, AUPR, FPR95, against SVHN for the convolutional models and the near and far OOD sets of \S\ref{sec:transformers} for the ViT experiments), and efficiency (parameters, multiply-adds). Robustness under input corruption is reported separately as graded degradation on CIFAR-100-C and ImageNet-C \citep{hendrycks2019benchmarking}, severities 1--5. Per-experiment hyperparameters and implementation details are in Appendix~\ref{sec:app-impl}.

\begin{table}[t]
\caption{TorchUncertainty benchmark on CIFAR-10/100 (ResNet-50, WideResNet-28-10). $\sigma$N-Ens rows are the means over three seeds, with standard deviations in Table~\ref{tab:cnn-std}. Within each setting the best value is bolded while accounting for a $\sigma$ N-Ens standard deviation on ties. $\sigma$N-Ens FineTune is the pretrained model converted and briefly fine-tuned (Appendix~\ref{sec:app-pretrained}), its reference is Single. ResNet-18 results are in Appendix~\ref{sec:app-cnn}.}
\label{tab:main}
\begin{center}
\footnotesize
\setlength{\tabcolsep}{3pt}
\begin{tabular}{lllrrrrrrrr}
\toprule
Data & Arch & Method & Acc\,$\uparrow$ & NLL\,$\downarrow$ & ECE\,$\downarrow$ & AUPR\,$\uparrow$ & AUROC\,$\uparrow$ & FPR95\,$\downarrow$ & Params\,$\downarrow$ & MAdds\,$\downarrow$ \\
\midrule
C10 & RN50 & Single & 95.1 & 0.211 & 3.1 & 95.2 & 91.9 & 23.6 & 23.52 & 1.30 \\
 &  & \textbf{$\sigma$N-Ens FineTune} & 94.3 & 0.178 & 0.99 & 95.7 & 92.2 & 23.3 & 24.38 & 5.22 \\
 &  & BatchEnsemble & 93.9 & 0.255 & 3.3 & 94.7 & 91.3 & 20.1 & 23.63 & 5.19 \\
 &  & MIMO & 95.4 & 0.197 & 3.0 & 95.1 & 90.8 & 26.0 & 23.59 & 1.30 \\
 &  & Masksembles & 95.3 & 0.175 & 1.9 & 95.7 & 92.2 & 22.1 & 23.81 & 5.19 \\
 &  & FiLM-Ensemble & 95.9 & 0.139 & 1.1 & 96.1 & 92.9 & 22.6 & 23.87 & 5.21 \\
 &  & Packed & 95.9 & 0.137 & \textbf{0.8} & \textbf{97.3} & \textbf{95.2} & \textbf{14.4} & 14.55 & 1.00 \\
 &  & Deep Ensembles & \textbf{96.0} & \textbf{0.136} & \textbf{0.8} & 97.0 & 94.7 & 15.5 & 94.08 & 5.19 \\
 &  & \textbf{$\sigma$N-Ens} & 95.4 & 0.156 & 1.2 & 96.5 & 93.7 & 19.8 & 24.38 & 5.22 \\
\cmidrule(l){2-11}
 & WRN & Single & 95.4 & 0.200 & 2.9 & 96.1 & 93.2 & 20.4 & 36.49 & 5.95 \\
 &  & \textbf{$\sigma$N-Ens FineTune} & 95.1 & 0.157 & 0.79 & 94.5 & 91.6 & 21.2 & 36.76 & 23.83 \\
 &  & BatchEnsemble & 95.6 & 0.206 & 2.7 & 95.5 & 92.5 & 22.1 & 36.59 & 23.81 \\
 &  & MIMO & 94.7 & 0.234 & 3.4 & 94.9 & 90.6 & 30.9 & 36.51 & 5.96 \\
 &  & Masksembles & 94.0 & 0.186 & 1.6 & 97.2 & 95.0 & 14.5 & 36.53 & 23.82 \\
 &  & FiLM-Ensemble & 95.9 & \textbf{0.127} & \textbf{0.6} & 98.0 & 96.0 & 13.0 & 36.60 & 23.82 \\
 &  & Packed & \textbf{96.2} & 0.133 & 0.9 & \textbf{98.1} & \textbf{96.5} & \textbf{11.1} & 19.35 & 4.06 \\
 &  & Deep Ensembles & 95.8 & 0.143 & 1.3 & 97.8 & 96.0 & 12.5 & 145.96 & 23.82 \\
 &  & \textbf{$\sigma$N-Ens} & 96.0 & \textbf{0.127} & 0.8 & 96.3 & 92.9 & 23.3 & 36.76 & 23.83 \\
\midrule
C100 & RN50 & Single & 78.3 & 0.905 & 8.9 & 87.4 & 77.9 & 57.6 & 23.70 & 1.30 \\
 &  & \textbf{$\sigma$N-Ens FineTune} & 77.5 & 0.788 & 1.94 & 87.1 & 76.0 & 63.8 & 24.38 & 5.22 \\
 &  & BatchEnsemble & 66.6 & 1.788 & 18.2 & 85.2 & 74.6 & 60.6 & 23.81 & 5.19 \\
 &  & MIMO & 79.0 & 0.876 & 7.9 & 87.5 & 76.9 & 64.7 & 24.33 & 1.30 \\
 &  & Masksembles & 78.5 & 0.832 & 4.6 & \textbf{90.3} & \textbf{81.9} & 52.3 & 23.81 & 5.19 \\
 &  & FiLM-Ensemble & 80.6 & 0.733 & 2.9 & 89.6 & 81.3 & \textbf{50.7} & 23.87 & 5.21 \\
 &  & Packed & \textbf{81.2} & \textbf{0.703} & 2.0 & 90.0 & 81.7 & 56.5 & 15.55 & 1.00 \\
 &  & Deep Ensembles & 80.9 & 0.713 & 2.6 & 89.2 & 80.8 & 52.5 & 94.82 & 5.19 \\
 &  & \textbf{$\sigma$N-Ens} & 80.8 & \textbf{0.701} & \textbf{1.6} & 89.0 & 80.7 & 52.4 & 24.38 & 5.22 \\
\cmidrule(l){2-11}
 & WRN & Single & 81.2 & 0.736 & 7.8 & 89.3 & 80.4 & 52.0 & 36.55 & 5.95 \\
 &  & \textbf{$\sigma$N-Ens FineTune} & 80.2 & 0.663 & \textbf{1.31} & 86.5 & 75.7 & 62.3 & 36.76 & 23.83 \\
 &  & BatchEnsemble & 82.3 & 0.835 & 13.0 & 88.1 & 78.2 & 69.8 & 36.65 & 23.81 \\
 &  & MIMO & 80.2 & 0.822 & 2.8 & 84.9 & 72.0 & 72.8 & 36.74 & 5.96 \\
 &  & Masksembles & 74.4 & 0.937 & 6.3 & 76.1 & 60.0 & 75.1 & 36.59 & 23.82 \\
 &  & FiLM-Ensemble & 82.8 & 0.618 & 3.9 & 89.8 & 81.4 & 55.3 & 36.60 & 23.82 \\
 &  & Packed & 83.9 & 0.678 & 8.9 & 86.2 & 73.2 & 80.7 & 36.62 & 5.95 \\
 &  & Deep Ensembles & 83.0 & 0.602 & 3.3 & \textbf{90.1} & \textbf{81.8} & \textbf{51.4} & 145.96 & 23.82 \\
 &  & \textbf{$\sigma$N-Ens} & \textbf{84.4} & \textbf{0.584} & 5.1 & \textbf{90.6} & \textbf{82.6} & 53.6 & 36.76 & 23.83 \\
\bottomrule
\end{tabular}
\end{center}
\end{table}

\subsection{Convolutional networks}\label{sec:cnn}
We compare against a single model, deep ensembles \citep{lakshminarayanan2017}, and the implicit ensemble baselines BatchEnsemble \citep{wen2020batchensemble}, MIMO \citep{havasi2021mimo}, Masksembles \citep{durasov2021masksembles}, Packed Ensembles \citep{laurent2022packed}, and FiLM-Ensemble \citep{turkoglu2022film}, all at $M=4$; baseline numbers are transcribed from \citet{laurent2022packed} and our rows run in the same protocol. We also reproduced the WideResNet/CIFAR-100 Single and Deep Ensembles rows, as the transcribed values were anomalous. We report two variants: $\sigma$N-Ens trained from scratch, and $\sigma$N-Ens FineTune, which converts the Single-model checkpoint and fine-tunes briefly. We use $\tau=0.1$, $\lambda=10^{-2}$ for all from-scratch benchmarks. This single value is fixed a priori and applied across every architecture and dataset, not tuned per setting. A $\tau$ sweep shows that better values exist, but a single value keeps the benchmark simple. Further details are in Appendix~\ref{sec:app-impl}.

 The $\sigma$N-Ens FineTune rows fold the pretrained variant into the main comparison; the relevant baseline for them is the Single checkpoint, not the from-scratch ensembles. Further convolutional results, including the ResNet-18 backbone and standard deviations, are in Appendix~\ref{sec:app-cnn}.

\subsection{Transformers}\label{sec:transformers}
We wrap every LayerNorm in the backbone with \sln{} via the output-preserving conversion of Appendix~\ref{sec:app-pretrained}, a small extension of TS\sbn{} \citep{tssbn} that we introduce for the pretrained models, and regularise the FFN-block scales only. For \snens{}, $\tau$ requires a higher value in the pretrained regime, where too much diversity harms the inherited representation. To be comparable Single, Deep Ensembles, LoRA-Ensemble, and \snens{}, share one public timm ViT-B/16 backbone and a single training recipe per dataset, with the full settings in Appendix~\ref{sec:app-impl}. Packed and FiLM ensembles do not appear here, since neither can wrap the pretrained backbone this benchmark uses and rely on training from scratch. Hydra Ensembles is excluded as its ViT backbone and code are not publicly available. The near and far OOD sets follow the framework defaults. For CIFAR-100, near OOD is CIFAR-10 and TinyImageNet and far OOD is MNIST, SVHN, Textures, and Places365. For ImageNet, near OOD is SSB-Hard and NINCO and far OOD is iNaturalist, Textures, and OpenImage-O. We report in-distribution, OOD, and corruption-shift results for both datasets in \S\ref{sec:results}.

\begin{table}[t]
\caption{Text classification on SST-2 (BERT-base), following the Hydra Ensembles protocol \citep{gabetni2026hydra}. The Heads column gives each method's attention-head configuration, which Hydra restructures by duplicating, pruning, and merging heads; \snens{} shares all 12 across its $M=3$ members. On this near-saturated task accuracy is flat across methods; calibration and OOD detection are where they separate.}
\label{tab:sst2}
\begin{center}
\footnotesize
\setlength{\tabcolsep}{4pt}
\begin{tabular}{lrrrrrrrrr}
\toprule
Method & Heads & Acc\,$\uparrow$ & Brier\,$\downarrow$ & NLL\,$\downarrow$ & ECE\,$\downarrow$ & aECE\,$\downarrow$ & AUROC\,$\uparrow$ & FPR95\,$\downarrow$ & AUPR\,$\uparrow$ \\
\midrule
Single             & 12  & 92.55 & 0.12 & 0.27 & 0.05 & 0.04 & 70.16 & 70.62 & 81.93 \\
Deep Ensembles     & 12  & \textbf{93.00} & \textbf{0.11} & 0.24 & 0.04 & 0.04 & 74.81 & 62.69 & \textbf{84.90} \\
MC Dropout         & 12  & 92.55 & 0.13 & 0.31 & 0.05 & 0.04 & 72.23 & 67.36 & 81.96 \\
LoRA-Ens     & 12  & 92.89 & 0.12 & 0.28 & 0.05 & 0.04 & 70.83 & 68.70 & 82.27 \\
Hydra-Ens (Circ)   & 6$\times$3 & 92.55 & 0.12 & 0.24 & 0.04 & 0.04 & \textbf{77.60} & \textbf{55.06} & 84.16 \\
\textbf{\snens{}}            & 12 & 91.93 & 0.13 & \textbf{0.23} & 0.04 & 0.04 & 75.38 & 58.45 & 84.38 \\
\bottomrule
\end{tabular}
\end{center}
\end{table}

\begin{table}[t]
\caption{ViT-B/16 benchmark. All methods share public pretrained backbones and one recipe; best per column in bold. \snens{} is the best-calibrated model without the deep ensemble's parameter cost, but weaker at out-of-distribution detection.}
\label{tab:vit}
\begin{center}
\footnotesize
\setlength{\tabcolsep}{3pt}
\begin{tabular}{llrrrrrrrr}
\toprule
& & & & & & \multicolumn{2}{c}{OOD AUROC\,$\uparrow$} & & \\
\cmidrule(lr){7-8}
& Method & Acc\,$\uparrow$ & NLL\,$\downarrow$ & ECE\,$\downarrow$ & Brier\,$\downarrow$ & near & far & Par.\,$\downarrow$ & MAdds\,$\downarrow$ \\
\midrule
\multirow{4}{*}{\rotatebox{90}{C100}}
& Single        & $92.77{\pm}0.10$ & $0.317{\pm}0.022$ & $4.23{\pm}0.54$ & $0.1185{\pm}0.0033$ & $92.75{\pm}0.24$ & $80.32{\pm}1.29$ & $\mathbf{85.9}$ & $\mathbf{16.9}$ \\
& Deep Ensembles & $\mathbf{93.12{\pm}0.00}$ & $0.261{\pm}0.008$ & $2.86{\pm}0.30$ & $0.1058{\pm}0.0012$ & $93.13{\pm}0.06$ & $81.60{\pm}0.74$ & $343.5$ & $67.4$ \\
& LoRA-Ens  & $92.09{\pm}0.12$ & $0.336{\pm}0.016$ & $7.87{\pm}1.51$ & $0.1276{\pm}0.0036$ & $\mathbf{93.22{\pm}0.21}$ & $\mathbf{85.69{\pm}0.51}$ & $88.5$ & $67.9$ \\
& \textbf{\snens{}} & $\mathbf{93.12{\pm}0.14}$ & $\mathbf{0.247{\pm}0.003}$ & $\mathbf{1.45{\pm}0.26}$ & $\mathbf{0.1055{\pm}0.0002}$ & $93.08{\pm}0.07$ & $80.25{\pm}1.12$ & $86.2$ & $67.4$ \\
\midrule
\multirow{4}{*}{\rotatebox{90}{IN-1K}}
& Single        & $82.12{\pm}0.09$ & $0.671{\pm}0.002$ & $4.11{\pm}0.08$ & $0.2566{\pm}0.0005$ & $79.04{\pm}0.26$ & $88.07{\pm}0.15$ & $\mathbf{86.6}$ & $\mathbf{16.9}$ \\
& Deep Ensembles & $\mathbf{83.67{\pm}0.02}$ & $\mathbf{0.593{\pm}0.001}$ & $1.34{\pm}0.11$ & $\mathbf{0.2347{\pm}0.0003}$ & $80.24{\pm}0.09$ & $90.22{\pm}0.04$ & $346.3$ & $67.4$ \\
& LoRA-Ens  & $83.27{\pm}0.15$ & $0.666{\pm}0.124$ & $6.14{\pm}9.21$ & $0.2518{\pm}0.0225$ & $\mathbf{80.93{\pm}0.35}$ & $\mathbf{91.84{\pm}0.14}$ & $91.2$ & $67.9$ \\
& \textbf{\snens{}} & $82.32{\pm}0.03$ & $0.655{\pm}0.005$ & $\mathbf{1.10{\pm}0.11}$ & $0.2526{\pm}0.0010$ & $77.95{\pm}0.09$ & $89.17{\pm}0.13$ & $88.9$ & $67.4$ \\
\bottomrule
\end{tabular}
\end{center}
\end{table}

\begin{figure}[t]
\begin{center}

\includegraphics[width=0.9\linewidth]{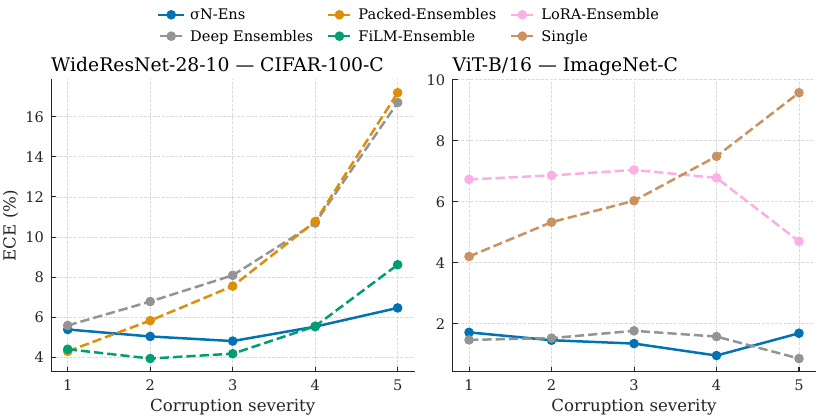}
\end{center}
\caption{Calibration under distribution shift: expected calibration error against corruption severity for a from-scratch WideResNet and a pretrained ViT-B/16. In both regimes \snens{}'s ECE stays nearly flat as corruption intensifies. Per-severity values for all backbones are in Appendix~\ref{sec:app-shift}.}
\label{fig:shift-main}
\vspace{-0.5cm}
\end{figure}

\subsection{Results}\label{sec:results}
Read together, the tables show that a single mechanism produces a calibrated ensemble across both convolutional and transformer architectures, whether trained from scratch or fine-tuned from a pretrained model.

On the convolutional benchmarks, our method is competitive across metrics, with low ECE and the best or near-best NLL in most settings. Its strongest single result is on WideResNet/CIFAR-100, where it achieves the best accuracy and NLL. Among related methods, Packed Ensembles is the most competitive, but it sits on the other branch of the implicit ensemble family. Its hard partition is fixed at construction, so it cannot add members without widening the architecture or reducing member capacity, exposes no accuracy-calibration control, and cannot wrap a pretrained backbone. Its default configuration of ${\alpha = 2, \gamma = 2}$ doubles the baseline network's width to preserve representational capacity while introducing an internal subgroup factor to enforce structural sparsity within members. The contrast is sharpest under member scaling, where it degrades while ours improves (\S\ref{sec:mscaling}). Appendix~\ref{sec:app-baselines} sets out the full comparison, and Appendix~\ref{sec:app-cost} reports the full compute and memory cost.

On transformers, the ensemble is obtained by wrapping the normalisation layers of a pretrained checkpoint and fine-tuning briefly. It does not require complex architectural modifications like Hydra Ensembles, and it has a lower parameter cost than LoRA-Ensemble (Appendix~\ref{sec:app-baselines}). Calibration is where the converted models stand out, most clearly under distribution shift. Figure~\ref{fig:shift-main} shows it holding steady in both regimes as corruption intensifies. On SST-2, accuracy separates weakly, and unlike the vision benchmarks, OOD detection is competitive.

Because every filter remains available to every member, added capacity is pooled rather than partitioned, so the ensemble can share features where capacity is scarce and specialise where it is abundant. Partitioning and fixed-rank methods cannot reallocate capacity this way, which the member-scaling ablation (\S\ref{sec:mscaling}) shows.

Out-of-distribution detection is where the shared backbone shows its limitation, and the two regimes manifest it differently, as predicted in \S\ref{sec:modunc}. Trained from scratch, members shape the representation under the diversity pressure, and detection holds up with capacity and difficulty. On the pretrained vision backbones, the representation is inherited, and members can only reweight what it already encodes, with limited success. The pretrained comparison also carries a caveat, since no method in these tables draws members from truly independent models, and all of them are derived from one checkpoint (\S\ref{sec:transformers}). Within this regime, detection tracks how far each method perturbs the inherited solution. LoRA-Ens, whose high-variance adapters perturb it most, leads on both datasets, while our modulation perturbs it least by design, and detection deteriorates.

\section{Ablations}\label{sec:ablations}
The two studies that validate the diversity mechanism, the accuracy-calibration tradeoff and the \sn{}-cosine to Jensen-gap correspondence, appear in \S\ref{sec:tradeoff} as the empirical anchors of the bias-variance-diversity framework. This section examines the method's two remaining degrees of freedom, the ensemble size $M$ and the regularisation strength $\lambda$.

\subsection{Scaling with member count}\label{sec:mscaling}
A property that distinguishes soft sharing from hard partitioning is how each behaves as the number of members grows. We compare \snens{} against Packed Ensembles, the archetypal partitioning method, on CIFAR-100 / ResNet-50 across $M$ from 2 to 16, holding $\tau = 0.5$ fixed, a neutral moderate value, rather than tuning it per $M$. The two methods sit at opposite ends of the $M$-choose-$k$ continuum of \S\ref{sec:mechanism}, and the sweep shows where each design assumption breaks down.

\begin{figure}[t]
\begin{center}
\includegraphics[width=0.9\linewidth]{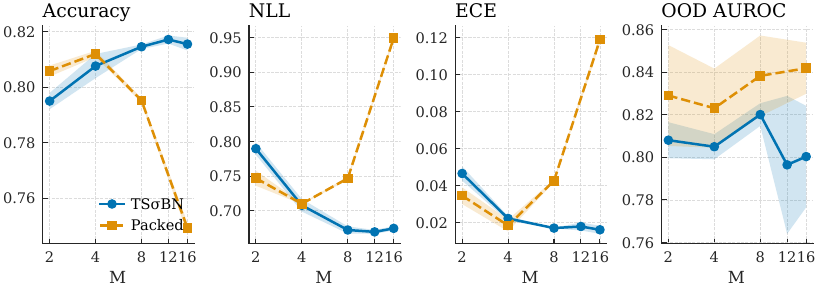}
\end{center}
\caption{Member scaling on CIFAR-100 / ResNet-50 at $\tau = 0.5$. \snens{} improves and then saturates as $M$ grows, while Packed degrades and collapses at the largest member counts; shaded bands are seed spread.}
\label{fig:mscaling}
\end{figure}

Figure~\ref{fig:mscaling} shows the crossover of performance, with exact numbers in Appendix~\ref{sec:app-mscaling}. At $M = 2$, Packed holds a small advantage, since its hard partition yields two independent sub-networks when capacity per member is still ample. By $M = 4$, the methods are at parity, and beyond it they diverge. \snens{} improves and then saturates, with its NLL flattening. Packed instead degrades and then collapses at the largest member counts, where forcing many disjoint channel groups through a fixed width leaves each member too thin to remain competitive. This mirrors the $M$-choose-$k$ reading. Packed is locked at $k = 1$, so per-member capacity falls as $M$ grows, whereas \snens{}'s soft allocation keeps filters shared where capacity is scarce. The claim here concerns scaling behaviour rather than absolute performance at small $M$, where Packed's structural independence is a real if narrow advantage.

\subsection{Regularisation strength}\label{sec:lambda}
\S\ref{sec:mechanism} casts $\tau$ as the control over the equilibrium and $\lambda$ as the strength of the pull toward it. If, as we argue, the from-scratch endpoint is governed primarily by $\tau$, then final-state performance should be largely insensitive to $\lambda$, leaving $\tau$ as the only hyperparameter a practitioner must set. We test this on from-scratch CNNs, sweeping $\lambda$ across three orders of magnitude at fixed $\tau$ and reporting converged metrics.

\begin{table}[t]
\caption{Regularisation-strength robustness, from-scratch CIFAR-100, single seed. $\lambda$ swept at fixed $\tau$; converged Acc / NLL / ECE.}
\label{tab:lambda}
\begin{center}
\small
\begin{tabular}{llcccc}
\toprule
Arch & $\tau$ & $\lambda = 10^{-4}$ & $\lambda = 10^{-3}$ & $\lambda = 10^{-2}$ & $\lambda = 10^{-1}$ \\
\midrule
ResNet-18 & 0.1 & 77.7 / 0.830 / 1.7 & 77.9 / 0.825 / 1.7 & 77.5 / 0.833 / 1.3 & 78.0 / 0.826 / 1.8 \\
ResNet-50 & 0.1 & 80.8 / 0.706 / 1.7 & 80.6 / 0.703 / 1.5 & 80.7 / 0.710 / 1.9 & 80.9 / 0.701 / 2.3 \\
WideResNet & 0.1 & 84.3 / 0.590 / 5.1 & 84.3 / 0.592 / 5.3 & 84.3 / 0.590 / 5.2 & 84.2 / 0.591 / 5.0 \\
\bottomrule
\end{tabular}
\end{center}
\end{table}

Table~\ref{tab:lambda} shows that accuracy and NLL are stable across the full $\lambda$ range on all three backbones, varying by less than a point in accuracy and a few thousandths in NLL as $\lambda$ moves over three orders of magnitude. This is the behaviour the $\tau$/$\lambda$ account predicts. Trained to convergence, the $\tau$-defined attractor governs the from-scratch endpoint, and the strength applied along the way leaves final performance essentially unchanged for any sufficiently non-trivial $\lambda$. We fix $\lambda = 10^{-2}$ for all from-scratch experiments and treat $\tau$ as the only user-facing control. In practice $\lambda$ need not be tuned per setting. It can be frozen at its default, leaving $\tau$ as the single control a practitioner adjusts.

Together with the member-scaling result and the temperature- and diversity-metric studies in \S\ref{sec:tradeoff}, this completes the empirical analysis. The method scales with ensemble size, and its default regularisation weight requires no tuning.

\section{Conclusion}\label{sec:conclusion}
We have presented a normalisation-based implicit ensemble in which a single temperature sets member diversity. Treating each member as a task in a multi-task architecture, a softmax regulariser drives the sigmoid-bounded normalisation scales toward a defined family of equilibria. Member disagreement is controlled during training rather than being a byproduct of initialisation or architecture. We show that this parameter-space diversity correlates with output diversity and traces the accuracy-calibration tradeoff. Furthermore, because only the normalisation layers are affected, the same mechanism applies across convolutional and transformer backbones and across training regimes, and can turn an existing pretrained model into a calibrated ensemble.

The method's principal limitation is a consequence of the same design that makes it efficient. Members share a single backbone and do not have their own feature extractors; they differ only in how they weight the features the backbone produces. The epistemic uncertainty they express, which we call modulation uncertainty, is therefore one component of what a full ensemble draws on, namely the disagreement reachable by reweighting shared features. The disagreement that requires differing feature extractors is structurally forgone, which is why the method is weaker on out-of-distribution detection, most clearly on pretrained backbones where the representation is inherited. The same structure makes it robust to distribution shift, where calibration degrades slowly as corruption intensifies. Giving members some feature-extraction capacity of their own, for instance through partially shared LoRA adapters, is the natural way to recover the forgone component, and we leave it to future work.

Empirically, the method is competitive with state-of-the-art approaches while using fewer parameters than a deep ensemble, and can continue to scale as members are added. A single temperature provides this control without complex architectural redesign, making controllable diversity regularisation of practical interest.

\bibliography{main}
\bibliographystyle{tmlr}

\appendix
\section{Appendix}\label{sec:appendix}

\subsection{Additional convolutional results}\label{sec:app-cnn}
Table~\ref{tab:rn18} reports the ResNet-18 backbone on CIFAR-10/100, extracted from the main benchmark for space. The reading matches the higher-capacity backbones: $\sigma$N-Ens is competitive across metrics, and the $\sigma$N-Ens FineTune row improves likelihood and calibration over the Single checkpoint it starts from.

\begin{table}[h]
\caption{ResNet-18 on CIFAR-10/100. Conventions as in Table~\ref{tab:main}; our method rows are highlighted, and per-seed std for $\sigma$N-Ens is in Table~\ref{tab:cnn-std}. $\sigma$N-Ens FineTune Params/MAdds match $\sigma$N-Ens.}
\label{tab:rn18}
\begin{center}
\footnotesize
\setlength{\tabcolsep}{3pt}
\begin{tabular}{lllrrrrrrrr}
\toprule
Data & Arch & Method & Acc\,$\uparrow$ & NLL\,$\downarrow$ & ECE\,$\downarrow$ & AUPR\,$\uparrow$ & AUROC\,$\uparrow$ & FPR95\,$\downarrow$ & Params\,$\downarrow$ & MAdds\,$\downarrow$ \\
\midrule
C10 & RN18 & Single          & 94.0 & 0.238 & 3.5 & 94.0 & 89.7 & 33.8 & 11.17 & 0.56 \\
    &      & \textbf{$\sigma$N-Ens FineTune} & 94.1 & 0.180 & 0.85 & 95.8 & 92.9 & 19.4 & 11.39 & 2.23 \\
    &      & BatchEnsemble   & 92.9 & 0.257 & 3.1 & 92.4 & 87.8 & 32.1 & 11.21 & 2.22 \\
    &      & MIMO            & 94.0 & 0.228 & 3.3 & 94.4 & 90.2 & 28.6 & 11.19 & 0.56 \\
    &      & Masksembles     & 94.0 & 0.188 & 0.9 & 93.6 & 89.5 & 27.8 & 11.24 & 2.22 \\
    &      & FiLM-Ensemble   & 94.1 & 0.180 & 0.7 & 94.3 & 90.6 & 25.7 & 11.25 & 2.22   \\
    &      & Packed          & 94.3 & 0.178 & 0.7 & 94.7 & 91.3 & 23.2 & 8.18  & 0.48 \\
    &      & Deep Ensembles  & 95.1 & 0.156 & 0.8 & 94.7 & 91.3 & 18.0 & 44.70 & 2.22 \\
    &      & \textbf{$\sigma$N-Ens}   & 93.9 & 0.185 & 0.7 & 93.4 & 89.1 & 30.6 & 11.39 & 2.23 \\
\midrule
C100 & RN18 & Single         & 75.1 & 1.016 & 9.3 & 88.6 & 79.5 & 55.0 & 11.22 & 0.56 \\
     &      & \textbf{$\sigma$N-Ens FineTune} & 76.9 & 0.831 & 2.09 & 86.7 & 75.9 & 62.5 & 11.39 & 2.23 \\
     &      & BatchEnsemble  & 71.2 & 1.236 & 11.6 & 86.0 & 75.4 & 60.2 & 11.25 & 2.22 \\
     &      & MIMO           & 75.3 & 0.962 & 6.9 & 89.2 & 80.7 & 52.9 & 11.36 & 0.56 \\
     &      & Masksembles    & 74.2 & 1.054 & 6.1 & 86.7 & 76.3 & 59.8 & 11.24 & 2.22 \\
     &      & FiLM-Ensemble  & 77.0 & 0.886 & 2.3 & 89.2 & 80.1 & 56.6 & 11.25 & 2.22   \\
     &      & Packed         & 76.4 & 0.858 & 4.1 & 88.7 & 79.8 & 57.1 & 8.27  & 0.48 \\
     &      & Deep Ensembles & 78.2 & 0.800 & 1.8 & 90.2 & 82.4 & 50.5 & 44.88 & 2.22 \\
     &      & \textbf{$\sigma$N-Ens}  & 77.8 & 0.826 & 1.6 & 86.8 & 76.5 & 61.5 & 11.39 & 2.23 \\
\bottomrule
\end{tabular}
\end{center}
\end{table}

Table~\ref{tab:cnn-std} reports the per-seed variability of $\sigma$N-Ens (from scratch) behind Tables~\ref{tab:main} and~\ref{tab:rn18}, as three-seed mean\,$\pm$\,std.

\begin{table}[h]
\caption{Per-seed variability of $\sigma$N-Ens (from scratch) on CIFAR-10/100, three-seed mean\,$\pm$\,std. These are the seed spreads referenced by Tables~\ref{tab:main} and~\ref{tab:rn18}; the transcribed baselines carry none.}
\label{tab:cnn-std}
\begin{center}
\footnotesize
\setlength{\tabcolsep}{5pt}
\begin{tabular}{llrrrrrr}
\toprule
Data & Arch & Acc\,$\uparrow$ & NLL\,$\downarrow$ & ECE\,$\downarrow$ & AUPR\,$\uparrow$ & AUROC\,$\uparrow$ & FPR95\,$\downarrow$ \\
\midrule
C10 & RN18 & $93.9{\pm}0.0$ & $0.185{\pm}0.002$ & $0.7{\pm}0.1$ & $93.4{\pm}1.1$ & $89.1{\pm}2.0$ & $30.6{\pm}5.3$ \\
 & RN50 & $95.4{\pm}0.0$ & $0.156{\pm}0.002$ & $1.2{\pm}0.0$ & $96.5{\pm}0.6$ & $93.7{\pm}0.9$ & $19.8{\pm}1.6$ \\
 & WRN & $96.0{\pm}0.1$ & $0.127{\pm}0.000$ & $0.8{\pm}0.2$ & $96.3{\pm}0.7$ & $92.9{\pm}1.4$ & $23.3{\pm}4.9$ \\
\midrule
C100 & RN18 & $77.8{\pm}0.3$ & $0.826{\pm}0.004$ & $1.6{\pm}0.3$ & $86.8{\pm}1.5$ & $76.5{\pm}2.9$ & $61.5{\pm}4.7$ \\
 & RN50 & $80.8{\pm}0.2$ & $0.701{\pm}0.005$ & $1.6{\pm}0.3$ & $89.0{\pm}0.3$ & $80.7{\pm}0.3$ & $52.4{\pm}1.7$ \\
 & WRN & $84.4{\pm}0.3$ & $0.584{\pm}0.004$ & $5.1{\pm}0.4$ & $90.6{\pm}0.7$ & $82.6{\pm}1.2$ & $53.6{\pm}3.8$ \\
\bottomrule
\end{tabular}
\end{center}
\end{table}

\subsection{Detailed comparison with competing baselines}\label{sec:app-baselines}
Several baselines match or beat \snens{} on an individual axis. We set out where each stands relative to the modulation approach, beginning with the most directly comparable method.

\paragraph{Packed Ensembles.} Packed Ensembles \citep{laurent2022packed} hard-partitions a single architecture into $M$ sub-models that share no weights, each given a slice of the channel width. The members are therefore fully independent. Unlike a modulation ensemble, Packed does not share or reshape a common representation; it divides capacity. It is best understood as an efficiency construct, a way to fit several independent networks into a single model's parameter and compute budget, rather than a statement about how members should differ.

Its strengths follow from that independence. Fully independent members give strong disagreement and, with it, a strong epistemic and out-of-distribution signal; and when capacity comfortably exceeds task difficulty, each slice is still wide enough to be a competent model, so the ensemble is at once accurate and cheap. Our results show this on the easier settings. On CIFAR-10, Packed has the best OOD detection in the table and high accuracy, among the lowest multiply-adds in the table.

The weaknesses follow from the same design. First, diversity is fixed at construction, and the partition exposes no accuracy-calibration control, so there is no analogue of sweeping $\tau$ along the tradeoff. Second, the architecture must be built partitioned, so Packed cannot wrap an existing pretrained backbone, the adaptation route central to our method. Third, adding members subdivides a fixed width. It reduces the capacity each receives, tolerable when the width greatly exceeds the task difficulty, but increasingly harmful as the difficulty rises. Each slice becomes too thin to be handled individually. This last point is visible in our own numbers. On the hardest, most capacity-constrained cell, CIFAR-100 on WideResNet, Packed's OOD detection collapses (AUROC $73.2$, FPR95 $80.7$). Its calibration degrades (ECE $8.9$), while \snens{} holds strong detection (AUROC $82.6$) and the best accuracy and NLL in the block; and under member scaling (\S\ref{sec:mscaling}), Packed degrades as $M$ grows, where the shared backbone lets our members improve.

We therefore read Packed less as a competitor on the same axis than as a different paradigm, partitioning independent sub-models versus modulating a shared one. Packed wins efficiency and, where capacity is ample, raw disagreement; the modulation approach wins controllable diversity, adaptation of pretrained backbones, and graceful member scaling, and overtakes Packed even on accuracy once capacity must be committed to a hard task. In the terms of \S\ref{sec:mechanism}, the two sit at opposite ends of the same axis. Packed is the $k=1$ hard partition built in by construction, while \snens{} reaches that regime only as a limit and can move off it under $\tau$.

\paragraph{LoRA-Ensemble.} LoRA-Ensemble \citep{muhlematter2025lora} shares our high-level aim, an implicit ensemble on a frozen pretrained transformer, and reaches it differently. Each member adds low-rank adapters to the attention projections, and the members differ only through the random initialisation of those adapters. Under its own protocol, it reports strong accuracy and calibration and is parameter-efficient. The difference that matters for us is how diversity is controlled. LoRA-Ensemble's diversity is governed by the adapter initialisation scale, a choice fixed before training; ours is governed by the temperature $\tau$, which moves the operating point along the accuracy-calibration tradeoff in a single step. That distinction is invisible when everything is tuned for one configuration and decisive when the configuration changes. An initialisation set for one recipe can over- or under-shoot under another, with no direct control to correct it, whereas $\tau$ adjusts the operating point directly.

For a controlled evaluation, we run every method under a shared protocol, adopted from Hydra \citep{gabetni2026hydra} and applied identically to each method, rather than tuning each separately, which is infeasible at this scale. This protocol differs from LoRA-Ensemble's own configuration in its optimisation scheme and ensemble size ($M=4$ here, versus $M=16$ in their paper). We report LoRA-Ensemble as it behaves under this common protocol; its results under its own tuned configuration are those of \citet{muhlematter2025lora}.

\paragraph{FiLM-Ensemble.} FiLM-Ensemble \citep{turkoglu2022film} is the prior method closest to ours in construction: an implicit ensemble that modulates a shared backbone through per-member feature-wise affine transformations. The two agree on the setting and differ in how member diversity is produced, how the modulation is parameterised, and how it is optimised. The most important difference is the source and control of diversity. FiLM-Ensemble seeds diversity by scaling the per-member affine initialisation, then trains every member on the same loss with no mechanism to maintain the spread, so the diversity of the trained model may drift as training proceeds. Our softmax regulariser instead drives the per-member importances toward a $\tau$-defined equilibrium, so $\tau$ sets where members converge and selects the operating point on the accuracy-calibration tradeoff. A consequence of the initialisation-seeded design is that FiLM-Ensemble is confined to training from scratch: a pretrained checkpoint's parameters are already set, leaving no initialisation to perturb, whereas our output-preserving conversion (Appendix~\ref{sec:app-pretrained}) lets the regulariser introduce diversity during a short fine-tune.

The modulation itself and its optimisation also differ. FiLM applies the full affine transformation per member, while \sn{} drops the shift and passes the scale through a sigmoid, yielding a bounded $[0,1]$ importance. The bound makes the member importances comparable, so the \sncos{} diversity metric is well conditioned and the equilibrium of \S\ref{sec:mechanism} is well defined; an unbounded affine supports neither reading. The optimisation also differs in a way that matters for diversity. FiLM-Ensemble trains all parameters under a single learning rate, whereas we apply a $100\times$ multiplier to the \sn{} scales (Appendix~\ref{sec:app-impl}) as per \cite{tssbn}.

\subsection{Compute and memory cost}\label{sec:app-cost}
We measure parameters, storage, multiply-adds, inference throughput, and peak memory on CIFAR-100 at $M=4$, in the mixed precision used throughout the paper and on a single A100 (Table~\ref{tab:cost}).

\begin{table}[h!]
\caption{Compute and memory cost on CIFAR-100 at $M=4$ on a single A100. The same settings are used as in the benchmarks: precision 16-mixed, batch size 128 for CNNs and 256 for ViT. }
\label{tab:cost}
\begin{center}
\footnotesize
\setlength{\tabcolsep}{4pt}
\begin{tabular}{ll rrr rr rr}
\toprule
& & \multicolumn{3}{c}{Static} & \multicolumn{2}{c}{Inference} & \multicolumn{2}{c}{Training} \\
\cmidrule(lr){3-5}\cmidrule(lr){6-7}\cmidrule(lr){8-9}
Arch & Method & Params & Storage & MAdds & img/s\,$\uparrow$ & VRAM & VRAM & Time \\
 & & (M) & (MB) & (G) & & (MB) & (MB) & (h) \\
\midrule
\multirow{7}{*}{RN50} & Single & 23.7 & 94.8 & 1.30 & 12622 & 410 & 2345 & 0.60 \\
 & BatchEnsemble & 23.9 & 95.6 & 5.22 & 1516 & 2526 & 22452 & 5.24 \\
 & MIMO & 24.3 & 97.3 & 1.31 & 12536 & 426 & 2359 & 0.66 \\
 & Masksembles & 23.8 & 95.2 & 5.22 & 1418 & 2659 & 12374 & 4.66 \\
 & Packed & 14.9 & 59.7 & 1.02 & 6368 & 669 & 3456 & 1.15 \\
 & Deep Ensembles & 94.8 & 379.3 & 5.22 & 3156 & 695 & 2345 & 2.42 \\
 & \textbf{\snens{}} & 24.4 & 97.5 & 5.22 & 3178 & 414 & 7250 & 2.83 \\
\midrule
\multirow{7}{*}{WRN} & Single & 36.5 & 146.2 & 5.96 & 8162 & 863 & 1836 & 1.01 \\
 & BatchEnsemble & 36.6 & 146.5 & 23.83 & 1437 & 3611 & 17687 & 5.96 \\
 & MIMO & 36.7 & 146.9 & 5.96 & 9225 & 877 & 1846 & 0.96 \\
 & Masksembles & 36.6 & 146.3 & 23.83 & 1372 & 3351 & 9404 & 5.52 \\
 & Packed & 36.6 & 146.5 & 5.96 & 5507 & 1555 & 3343 & 1.47 \\
 & Deep Ensembles & 146.2 & 584.8 & 23.83 & 2041 & 1302 & 1836 & 4.05 \\
 & \textbf{\snens{}} & 36.8 & 147.0 & 23.83 & 2057 & 864 & 5966 & 4.29 \\
\midrule
\multirow{4}{*}{ViT} & Single & 85.9 & 343.5 & 16.87 & 3371 & 1930 & 18585 & 0.64 \\
 & LoRA-Ens & 88.5 & 353.9 & 67.93 & 619 & 1766 & 58025 & 2.84 \\
 & Deep Ensembles & 343.5 & 1374.0 & 67.47 & 843 & 2960 & 18585 & 2.57 \\
 & \textbf{\snens{}} & 86.2 & 344.7 & 67.47 & 844 & 1936 & 70153 & 2.61 \\
\bottomrule
\end{tabular}
\end{center}
\end{table}

The compute costs split along how an ensemble produces its members. Deep ensembles pay mainly in parameters, storage and the number of training runs, since every member is an independent full model. Implicit ensembles avoid those costs by sharing a single set of weights and training in one run. Within the implicit family, partitioning methods cut compute by dividing the network among members, at the cost of each member's capacity. Modulation methods instead keep the full backbone for every member, so their parameters stay close to a single model's while their compute matches a deep ensemble's. What stands out is that multiply-adds alone are an incomplete measure: methods with the same MAdds differ noticeably in real speed and memory, depending on how much extra machinery their per-member modulation adds. Training all members jointly also carries a memory cost that a deep ensemble, trained one model at a time, avoids. \snens{} sits in the modulation family, at its low-overhead end: it keeps the parameter, storage, and single-run training advantages over deep ensembles, and matches deep ensembles on compute by design rather than competing with partitioning methods on it.

\subsection{Adapting pretrained backbones}\label{sec:app-pretrained}
The from-scratch construction of \S\ref{sec:arch} trains \sn{} in place of each normalisation layer. The same mechanism wraps a pretrained backbone, with one requirement. The conversion must leave the network's output unchanged, so fine-tuning starts from the pretrained function rather than a perturbed one.

The obstacle is that $\sigma(\gamma)\in(0,1)$ cannot reproduce a pretrained scale that lies outside it. We therefore reintroduce a per-layer scale magnitude $k_\ell$ and the per-channel shift $\beta_\ell$, used only to reconstruct the pretrained signal. For a member's trainable logit $\gamma$, the converted layer applies
\begin{equation}
y_\ell(\gamma) = \underbrace{k_\ell\,\sigma(\gamma)}_{\text{reconstructed scale}} \odot \operatorname{norm}_\ell(x) + \beta .
\end{equation}
As in the original affine, $\gamma$ and $\beta$ are per-channel and $\sigma$ acts channel-wise; the only per-layer quantity is the scalar $k_\ell = \max_c |\gamma_{\ell,c}^{\text{pre}}| / 0.95$, one value shared across the layer's channels (the $0.95$ keeps the largest scale off the saturation wall at $\sigma^{-1}(1)$). The pretrained bias is copied into $\beta$, and the logits are initialised per channel by $\gamma_{\ell,c}^{\text{new}} = \sigma^{-1}(\gamma_{\ell,c}^{\text{pre}} / k_\ell)$, so that $k_\ell\,\sigma(\gamma_{\ell,c}^{\text{new}}) = \gamma_{\ell,c}^{\text{pre}}$ for every channel and $y_\ell$ matches the pretrained output at initialisation.

Crucially, $k_\ell$ and $\beta$ are shared across members and frozen throughout training, so they reconstruct the pretrained signal without compensating for the diversity introduced by the regulariser. The regulariser acts on $\sigma(\gamma)$ alone, which stays the only member-specific, bounded quantity; the conversion is an output-preserving reparameterisation orthogonal to the diversity mechanism, and the equilibrium, the \sn{}-cosine metric, and $\tau$ as the sole control all carry over unchanged. The from-scratch construction is the special case $k_\ell=1,\ \beta=0$.

This wider scale matters mainly for transformers. Convolutional scales already lie in $(0,1)$, so $k_\ell=1$ suffices, and the layer reduces to plain \sbn{}. Transformer backbones instead show large magnitude differences between attention and MLP blocks, with LayerNorm scales growing with depth, so a bounded scale alone cannot reconstruct them. TS\sbn{} \citep{tssbn} handled this only partially, wrapping a selected subset of layers where the resulting error was acceptable; the per-layer scale here lets us wrap every normalisation layer, as long as the pretrained scales are positive, which they are in all backbones we tested. Member heads are reinitialised, since a shared head is a symmetric fixed point at which the regulariser has no gradient, so reinitialisation supplies the asymmetry the mechanism needs.

\subsection{Implementation details}\label{sec:app-impl}
We report only where we deviate from the protocols we build on; base settings are as in the cited references. Two choices are common to all experiments: the backbone \sn{} scales are trained with the $100\times$ learning-rate multiplier suggested by TS\sbn{} \citep{tssbn}, adopted unchanged rather than tuned, and no weight decay, and every member head places an additional normalisation layer (BN or LN) before its randomly initialised linear layer. This extra layer keeps the scale of the last shared backbone layer in line with the others, which the classifier that follows would otherwise inflate. The set of regularised layers, and the ViT temperatures, are selected on a held-out validation set.

\paragraph{Convolutional networks.} We follow the Packed Ensembles training protocol \citep{laurent2022packed} (SGD with a step schedule; learning rate, weight decay, and epochs vary by dataset and model as specified there), deviating only in setting dropout to zero. From-scratch \sbn{} scales are initialised from $\mathcal{N}(0,1)$, with $\tau=0.1$, $\lambda=10^{-2}$, and $M=4$. The diversity regulariser is applied to one \sn{} layer per residual block: the second normalisation layer in ResNet-18, the first in WideResNet-28-10, and the first and last in ResNet-50. The $\sigma$N-Ens FineTune variant converts a pretrained single-model checkpoint (Appendix~\ref{sec:app-pretrained}), reinitialises the heads, and fine-tunes with Adam at $10^{-3}$ for 25 epochs at $\tau=0.1$, $\lambda=10^{-4}$.

\paragraph{Transformers.} For the ViT experiments, all methods share one recipe: SGD (momentum $0.9$, no weight decay), learning rate $0.01$, $500$-step linear warmup then cosine decay, batch size $256$, no dropout, $M=4$, three seeds. Backbones are public timm ViT-B/16 \texttt{in21k} checkpoints, original-augmentation on CIFAR-100 and AugReg on ImageNet, with $10{,}000$ training steps on CIFAR-100 and $20{,}000$ on ImageNet. For \snens{}, scales are initialised by the output-preserving conversion of Appendix~\ref{sec:app-pretrained}, the regulariser acts on the FFN-block scales only, and we deviate from the shared recipe by using no warmup; we set $\lambda=10^{-3}$ throughout with $\tau=0.5$ on CIFAR-100 and $\tau=1.0$ on ImageNet, and $\tau=0.5$, $\lambda=10^{-3}$ on SST-2 / BERT-base, where we follow Hydra's protocol at $M=3$ with five-seed means rather than the $M=4$, three-seed ViT setting above. The LoRA-Ensemble baseline \citep{muhlematter2025lora} freezes the backbone and trains per-member rank-8 adapters on the three attention projections ($\alpha/r=1$, high-variance down-projection initialisation for diversity) plus a per-member head, under the same SGD recipe. Applying one shared recipe to all methods, rather than tuning each method separately, is a deliberate choice to enable a controlled comparison, as per-method grid search is infeasible at this scale. This recipe differs from LoRA-Ensemble's own configuration in both its optimisation scheme and its ensemble size, their paper uses $M=16$ members whereas we use $M=4$ throughout.

\paragraph{Choosing $\tau$.} In practice $\tau$ lives in a small range, roughly $[0.1, 1.0]$, and the tradeoff is smooth, so a few coarse values are enough to find a good operating point. A simple regime heuristic covers most cases: a low $\tau$ when training from scratch, where the filters are still forming, and a higher $\tau$ when adapting a pretrained backbone, where too much diversity harms the inherited representation. Because the induced diversity is readable from the weights through the \sn{}-cosine, the effect of a given $\tau$ can be checked without a full training and evaluation sweep.

\subsection{Temperature limits of the diversity penalty}\label{sec:app-limits}
We characterise the diversity penalty in isolation, as a function of the
importances alone. The penalty contributes a sharpening force whose strength
depends on $\tau$. The equilibria observed in real training are not fixed
points of the penalty alone. They arise where this force balances the task
loss.

The simulation of \S\ref{sec:mechanism} (Figure~\ref{fig:composite}~\figcenter{})
applies this penalty on its own, with no task loss. We optimise an $M \times C$
importance matrix with $C=512$, initialised from $\mathcal{N}(0,1)$ as in
training, using Adam at learning rate $0.1$ for $300$ steps, and report the mean
pairwise cosine once the member assignment stabilises, at around $150$ steps.
Feature ownership is read by thresholding $\sigma(\gamma)$ at $0.5$. Because the
sigmoid gradient vanishes at the extremes, each finite run settles at a soft
allocation rather than a hard corner, so the cosine surface varies smoothly
with $\tau$.

Fix one feature in one regularised layer and let $s_m = \sigma(\gamma_m) \in
(0,1)$ be member $m$'s importance for it, with across-member shares
$p = \softmaxt(s)$. The penalty for this feature is
$\ell = \sum_{m} \log p_m$, and its gradient is
\begin{equation}
\frac{\partial \ell}{\partial s_j} = \frac{1}{\tau}\big(1 - M p_j\big) .
\end{equation}
By concavity, $\ell$ is maximised exactly at the uniform distribution, so
descent moves away from equal sharing: it raises every member whose share is
above average ($p_j > 1/M$) and lowers the rest, at an overall rate of
$1/\tau$. The limits follow from the standard softmax behaviour. As
$\tau \to 0$ the shares approach one-hot at the leading member, and the
penalty drives each feature toward a single owner. As $\tau \to \infty$ the
shares flatten to $1/M$ regardless of $s$ and the gradient vanishes: the
regulariser stops contributing. It does not push members to be identical; it
leaves the configuration to the task loss and the initialisation.

The sigmoid keeps this well-posed. On unbounded logits, minimising $\ell$ is
degenerate: the leading logit grows without limit and the penalty has no
minimum. The bound $s_m \in (0,1)$ caps the logit spread at $1$, keeping the
penalty bounded below, and caps how sharp the realised softmax can become:
even with one importance at $1$ and the rest at $0$, the peak share is
$e^{1/\tau}\big/\big(e^{1/\tau} + M - 1\big)$. Moreover, since training descends on the logits $\gamma$, each
update carries the factor $\sigma'(\gamma_m) = s_m(1-s_m)$, which vanishes at
the extremes: allocations approach $0$ and $1$ asymptotically, so every
finite run ends at a soft allocation strictly inside the box.

\paragraph{The attractors are $k$-owner corners.} For $1 \le k < M$, call a configuration with $k$ members at
importance $1$ (the owners) and $M-k$ at importance $0$ a \emph{$k$-owner
corner}. Evaluating the shares at such a corner (softmax denominator
$k e^{1/\tau} + M - k$) places each owner above the average $1/M$ and each
non-owner below it, so the gradient pushes owners up and non-owners down: the
flow points into the corner, at every $k < M$ and every finite $\tau$. The
penalty's attractors are therefore the $k$-owner corners, the $M$-choose-$k$
allocation of \S\ref{sec:mechanism} derived rather than observed. The two
ends of the family bracket the designs of \S\ref{sec:relwork}: $k = 1$ is the
hard partition that partitioning methods build in by construction, and full
sharing lies outside the family, as the uniform state the penalty moves away
from.

\paragraph{How $\tau$ controls the allocation.} The temperature does not
change which corners exist; it changes the force driving features toward
them. Low $\tau$ pushes hard toward concentrated ownership, high $\tau$
pushes weakly, with the two limits above as endpoints. In training, each
feature settles where this $\tau$-shaped force balances the task loss, which
resists allocations that starve members of useful features. Raising $\tau$
weakens the push, so features settle nearer to sharing; lowering it
strengthens the push, so they settle nearer to single ownership. Which
members end up owning a given feature depends on its initial configuration
and the training dynamics; the simulation of \S\ref{sec:mechanism} maps the
resulting allocations empirically. This is the sense in which $\tau$ selects
the operating point, not by redefining the destinations but by setting how
far toward hard partitioning the ensemble is driven before the task loss
holds it.

\subsection{Modulation uncertainty: a component of the epistemic term}\label{sec:app-modunc}
We expand the decomposition of \S\ref{sec:modunc} and make the identification precise. Notation follows \S\ref{sec:modunc}: member $i$ predicts $q_i(y\mid x) = f(y\mid x, \thetaconv, \phi_i)$, with the backbone $\thetaconv$ shared across members and only the \sn{} importances $\phi_i$ varying with the member index.

\paragraph{A reachable-set view.} Fixing the trained backbone at $\thetaconv^\star$, every predictor our ensemble can instantiate lies in
\begin{equation}
\mathcal{F}_{\mathrm{mod}} = \big\{\, f(\cdot \mid \thetaconv^\star, \phi) : \phi \in \Phi \,\big\},
\qquad
\mathcal{F}_{\mathrm{DE}} = \big\{\, f(\cdot \mid \theta, \phi) : \theta \in \Theta,\ \phi \in \Phi \,\big\},
\end{equation}
where $\mathcal{F}_{\mathrm{DE}}$ is the family an ensemble with a free
backbone draws its members from. Since $\thetaconv^\star$ is one point of
$\Theta$, $\mathcal{F}_{\mathrm{mod}} \subseteq \mathcal{F}_{\mathrm{DE}}$,
strictly so whenever varying the backbone changes the predictor. Any
disagreement that requires two members to extract different features from the
same input is unreachable for us by construction; our members realise only the
disagreement obtainable by reweighting the features $\thetaconv^\star$ already
produces. The inclusion is deliberately elementary; its role is to fix
intuition for the decomposition.

\paragraph{The chain-rule decomposition.} To place the two ensembles in one
frame, consider the idealised ensemble in which both parts vary: a distribution
over pairs $(\theta, \phi)$, of which our method is the slice at
$\theta = \thetaconv^\star$. Its epistemic term splits by the chain rule of
mutual information,
\begin{equation}\label{eq:chain}
\MI{Y;(\theta,\phi)\mid x} \;=\;
\underbrace{\MI{Y;\theta\mid x}}_{\text{backbone component}}
\;+\;
\underbrace{\MI{Y;\phi\mid x,\theta}}_{\text{modulation component}} ,
\end{equation}
an identity requiring no independence assumptions. The first component is
disagreement attributable to varying the feature extractor itself, what
independently trained members express \citep{fort2019}; the second is
disagreement attributable to varying the modulation with the extractor held
fixed. Our ensemble realises the second component at the single draw
$\theta=\thetaconv^\star$,
\begin{equation}
\Umod(x) \;=\; \MI{Y;\phi \mid x,\ \theta=\thetaconv^\star} ,
\end{equation}
evaluated at our single trained backbone rather than averaged over $\theta$,
so it is one realisation of the conditional term rather than the conditional
term itself; the backbone component is forgone entirely. A deep ensemble varies the full
network and so draws on both components; the backbone component is the part of
its epistemic uncertainty that our shared-backbone construction cannot reach.

\paragraph{The absent component.} The decomposition is structural and says nothing, on its own, about how large each component is at a given input. Independently trained backbones converge to solutions that agree on in-distribution data and diverge on inputs far from it \citep{fort2019}. The backbone component $\MI{Y;\theta\mid x}$ is therefore small when $x$ is drawn from the distribution the backbone family represents, and large off-distribution, where it is precisely the signal a deep ensemble detects with. A corrupted input is a perturbation of in-distribution data the backbone still represents and little is lost by sharing. The modulation component still expresses calibrated disagreement, which we observe as a calibration that degrades slowly under corruption, holding ECE similar with deep ensembles across severities where a single model does not (\S\ref{sec:results}). At the out-of-distribution edge, the backbone component dominates the epistemic term, and it is exactly the part our members cannot express. OOD detection is correspondingly our weakest axis on both pretrained vision settings, near-OOD on ImageNet and far-OOD on CIFAR-100 (\S\ref{sec:results}), the predicted cost of sharing the backbone rather than an incidental shortfall. Members can still disagree somewhat on such inputs, through their reinitialised heads and reweightings of weak features, but not with the representation-level divergence that makes deep-ensemble disagreement informative there.

The same decomposition applies to any shared-backbone ensemble, including FiLM-Ensemble \citep{turkoglu2022film}: each expresses a modulation component and forgoes the backbone component. What the bounded \sn{} primitive adds is
not a different split but a controllable, measurable position within the component that is expressed, set through $\tau$ and monitored through the \sncos{} metric, where initialisation-seeded modulation leaves that position to drift.

\subsection{Diversity-metric correspondence}\label{sec:app-fourmetric}
Section~\ref{sec:tradeoff} validates \sn{}-cosine, a parameter-space metric, as a proxy for the output-space diversity that the bias-variance-diversity framework rewards. Here we give the full correspondence. The distinction is between \emph{parameter} diversity, how differently members weight shared features, which our regulariser acts on directly, and \emph{output} diversity, how differently members predict, which is what the decomposition contains. We compare four metrics:
\begin{itemize}
\item \textbf{\sn{}-cosine} (parameter-space, ours): the mean per-layer pairwise cosine between member importance vectors $\sigma(\boldsymbol\gamma)$, averaged over all normalisation layers. Computed from the weights alone, with no forward pass and no labels.
\item \textbf{Jensen gap} $\mathcal{J}$ (output-space, label-dependent) \citep{abe2022}: the Wood diversity term under the arithmetic-mean combiner \citep{wood2023bvd}, equal to the average member cross-entropy minus the cross-entropy of the averaged prediction.
\item $\MI{Y;M\mid x}$ (output-space, label-free): the Smith \& Gal \citep{smithgal2018} epistemic term, equal to the average KL divergence from each member to the arithmetic-mean prediction.
\item \textbf{Geometric ambiguity} (output-space, label-free): the ambiguity term under the cross-entropy's centroid combiner, the normalised geometric mean \citep{wood2023bvd}. It is empirically indistinguishable from $\MI{Y;M\mid x}$ in our setting ($|\rho|>0.998$ across all backbones), the centroid distinction falling below noise, so we report it only here.
\end{itemize}
Table~\ref{tab:fourmetric} reports the correlation of \sn{}-cosine against the three output-space metrics across a $\tau$ sweep (three seeds) on CIFAR-100. Correlations are strong throughout, with a floor of $|\rho|\ge0.83$ (Pearson) and $|\rho|\ge0.87$ (Spearman). Spearman exceeds Pearson on every setting, indicating a monotonic relationship with mild curvature, consistent with \sn{}-cosine saturating at the $\tau$ extremes while the output-space measures grow more linearly.

\begin{table}[h]
\caption{Correlation of \sn{}-cosine (all-layer average) with three output-space diversity metrics, across a $\tau$ sweep ($3$ seeds) on CIFAR-100. All values are negative (higher cosine, lower diversity).}
\label{tab:fourmetric}
\begin{center}
\small
\begin{tabular}{lrrrrrr}
\toprule
& \multicolumn{3}{c}{Pearson} & \multicolumn{3}{c}{Spearman} \\
\cmidrule(lr){2-4}\cmidrule(lr){5-7}
Backbone & $\MI{Y;M\mid x}$ & $\mathcal{J}$ & Geometric & $\MI{Y;M\mid x}$ & $\mathcal{J}$ & Geometric \\
\midrule
ResNet-18         & $-0.96$ & $-0.97$ & $-0.95$ & $-0.99$ & $-0.99$ & $-0.99$ \\
ResNet-50         & $-0.89$ & $-0.83$ & $-0.90$ & $-0.95$ & $-0.89$ & $-0.93$ \\
WideResNet-28-10  & $-0.86$ & $-0.85$ & $-0.86$ & $-0.88$ & $-0.87$ & $-0.87$ \\
\bottomrule
\end{tabular}
\end{center}
\end{table}

\subsection{Additional tradeoff curves and importance embeddings}\label{sec:app-frontiers}
\begin{figure}[h]
\begin{center}
\includegraphics[width=0.32\linewidth]{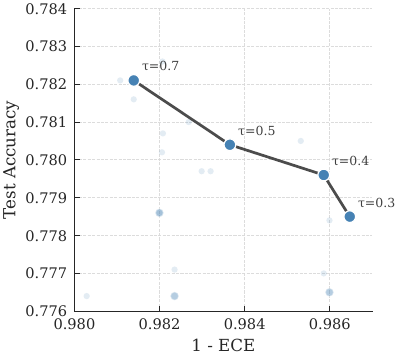}\hfill
\includegraphics[width=0.32\linewidth]{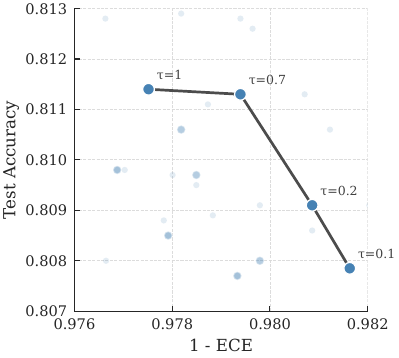}\hfill
\includegraphics[width=0.32\linewidth]{figures/pareto_c100wrn.pdf}
\end{center}
\caption{Accuracy-calibration tradeoff over a $\tau$ sweep on CIFAR-100 (3 seeds): ResNet-18 (left), ResNet-50 (centre), WideResNet-28-10 (right).}
\label{fig:app-frontiers}
\end{figure}

Figure~\ref{fig:app-frontiers} shows the accuracy-calibration tradeoff of \S\ref{sec:tradeoff} for all three convolutional backbones. Each shows monotone movement as $\tau$ is swept, with the WideResNet panel reproducing the main-text figure.

\begin{figure}[h]
\begin{center}\includegraphics[width=0.7\linewidth]{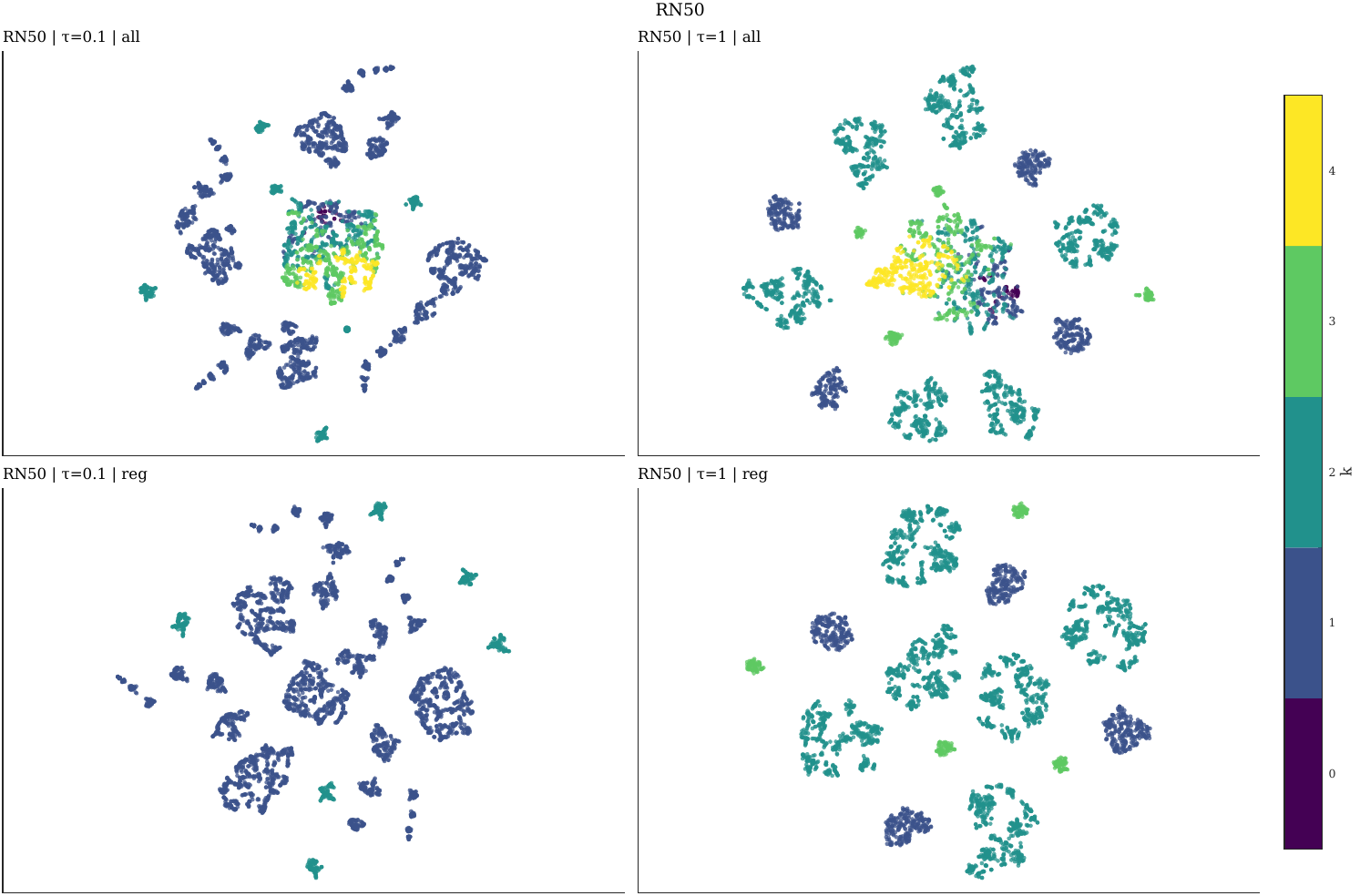}\end{center}
\caption{ResNet-50 importance embeddings; coloured by member-ownership $k$. Columns: $\tau=0.1$ (left), $\tau=1$ (right). Rows: all layers (top), regularised layers (bottom).}
\label{fig:embed-rn50}
\end{figure}

\begin{figure}[h]
\begin{center}\includegraphics[width=0.7\linewidth]{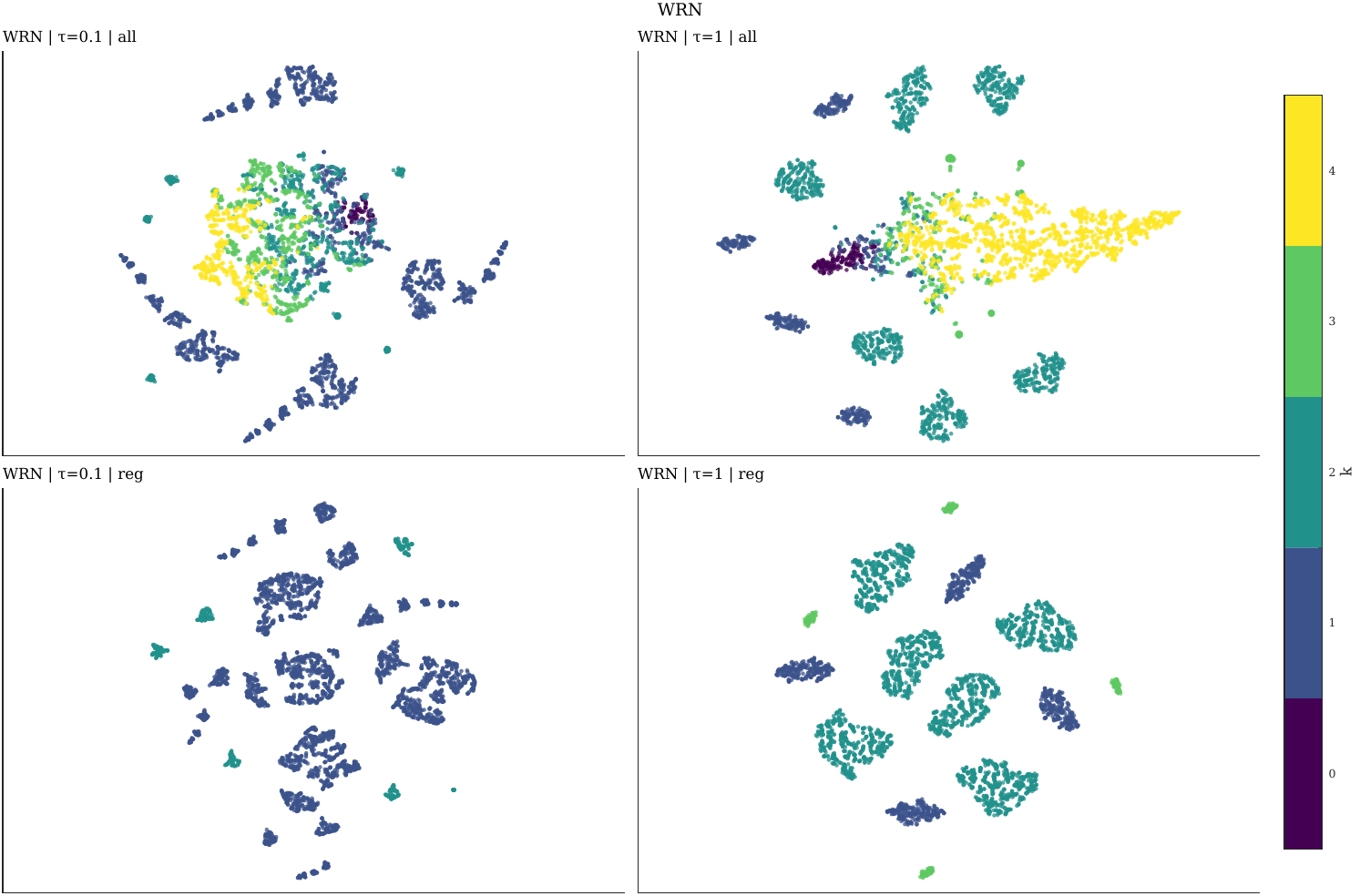}\end{center}
\caption{WideResNet-28-10 importance embeddings; coloured by member-ownership $k$. Columns: $\tau=0.1$ (left), $\tau=1$ (right). Rows: all layers (top), regularised layers (bottom).}
\label{fig:embed-wrn}
\end{figure}

Figures~\ref{fig:embed-rn50}--\ref{fig:embed-wrn} show the member-ownership structure directly. Each panel embeds the per-filter importance vectors and colours each filter by $k$, the number of members that place substantial importance on it ($k=1$ single-owner through $k=M$ fully shared). Within each figure, the left column is $\tau=0.1$ and the right is $\tau=1$; the top row aggregates all normalisation layers, and the bottom row aggregates only the regularised layers.

Two readings stand out and are consistent across architectures. First, $\tau$ shifts the ownership distribution as the mechanism predicts: on the regularised layers, $\tau=0.1$ drives most filters to $k=1$ (single-owner, the diverse extreme), while $\tau=1$ moves mass to $k=1,2$ with a few at $k=3$. Second, the regularised and unregularised layers play complementary roles: the soft partitioning induced by the regulariser on its target layers coexists with heavy sharing ($k=3,4$) on the untouched layers, as visible in the all-layer rows. The diverse member-specific allocation is thus concentrated where the regulariser acts, on a backbone that remains largely shared elsewhere.

\subsection{Member-scaling numbers}\label{sec:app-mscaling}
Table~\ref{tab:mscaling} reports the exact values underlying the member-scaling sweep of Figure~\ref{fig:mscaling}.

\begin{table}[t]
\caption{Member-scaling numbers behind Figure~\ref{fig:mscaling}, \snens{} against Packed Ensembles on CIFAR-100 / ResNet-50 as the member count $M$ grows ($\tau=0.5$ throughout); three-seed mean\,$\pm$\,std for both methods. \snens{} improves and then saturates, while Packed is strong at small $M$ but degrades once its hard partition makes each member too thin; $M=12$ is architecturally invalid for Packed under its grouping constraint. Packed's OOD AUROC instead rises with $M$: its fully independent members still disagree strongly on outliers even as their accuracy and calibration collapse, consistent with the independence account of Appendix~\ref{sec:app-baselines}.}
\label{tab:mscaling}
\begin{center}
\footnotesize
\setlength{\tabcolsep}{3pt}
\begin{tabular}{lrrrrrrrr}
\toprule
& \multicolumn{4}{c}{\snens{}} & \multicolumn{4}{c}{Packed} \\
\cmidrule(lr){2-5}\cmidrule(lr){6-9}
$M$ & Acc\,$\uparrow$ & NLL\,$\downarrow$ & ECE\,$\downarrow$ & AUROC\,$\uparrow$ & Acc\,$\uparrow$ & NLL\,$\downarrow$ & ECE\,$\downarrow$ & AUROC\,$\uparrow$ \\
\midrule
2  & $79.5{\pm}0.3$ & $0.790{\pm}0.009$ & $4.7{\pm}0.4$ & $80.8{\pm}0.8$ & $80.6{\pm}0.2$ & $0.747{\pm}0.011$ & $3.5{\pm}0.5$ & $82.9{\pm}2.4$ \\
4  & $80.8{\pm}0.4$ & $0.708{\pm}0.009$ & $2.2{\pm}0.1$ & $80.5{\pm}0.6$ & $81.2{\pm}0.1$ & $0.710{\pm}0.001$ & $1.9{\pm}0.3$ & $82.3{\pm}1.9$ \\
8  & $81.5{\pm}0.1$ & $0.672{\pm}0.005$ & $1.7{\pm}0.0$ & $82.0{\pm}0.5$ & $79.5{\pm}0.1$ & $0.747{\pm}0.002$ & $4.3{\pm}0.1$ & $83.8{\pm}1.9$ \\
12 & $81.7{\pm}0.1$ & $0.670{\pm}0.003$ & $1.8{\pm}0.2$ & $79.6{\pm}3.2$ & --             & --                & --            & --             \\
16 & $81.6{\pm}0.2$ & $0.675{\pm}0.003$ & $1.6{\pm}0.2$ & $80.0{\pm}2.4$ & $74.9{\pm}0.2$ & $0.949{\pm}0.004$ & $11.9{\pm}0.2$ & $84.2{\pm}1.2$ \\
\bottomrule
\end{tabular}
\end{center}
\end{table}

\subsection{Distribution-shift protocol and additional backbones}\label{sec:app-shift}
We evaluate robustness on the corruption benchmarks CIFAR-100-C and ImageNet-C \citep{hendrycks2019benchmarking}, which apply 15 synthetic corruptions (noise, blur, weather, and digital types) at five severity levels to the clean test sets. For each severity level, we average all metrics across the 15 corruption types and report accuracy, NLL, and ECE as a function of severity. Models are those of \S\ref{sec:results}, evaluated without any corruption-specific adaptation; the corrupted sets are used only at test time. Figure~\ref{fig:shift-main} shows the ImageNet ViT-B/16 result in the main text; per-severity numbers are collected in Tables~\ref{tab:shift-cnn} (from-scratch CNNs) and~\ref{tab:shift-vit} (pretrained ViT-B/16). On the from-scratch WideResNet, \snens{} achieves the best accuracy and NLL across all severities and the flattest ECE, holding near $5$--$6$\%, while deep ensembles and Packed surpass $16$\%. On the lower-capacity ResNet-50, the advantage narrows, as the capacity trend of \S\ref{sec:mscaling} predicts: \snens{} stays competitive and second-best on ECE at every severity. At the same time, Packed takes calibration, and deep ensembles take accuracy. On the pretrained ViT-B/16, \snens{} achieves ECE on par with a deep ensemble on ImageNet-C (both $\approx$1--2\%, with the single model degrading from 4 to 10\%) and attains the best NLL and ECE across the lower and middle corruption severities on CIFAR-100-C.

\begin{table}[ht]
\caption{Distribution shift on the from-scratch convolutional backbones (CIFAR-100-C): per-severity accuracy, NLL, and ECE, each aggregated over the 15 corruption types ($M=4$, three seeds). Acc and ECE in \%; bold marks the best value per column within each setup. Interpretation in the text above.}
\label{tab:shift-cnn}
\begin{center}
\small
\setlength{\tabcolsep}{5pt}
\begin{tabular}{ll l rrrrr}
\toprule
Setup & Method & Metric & s1 & s2 & s3 & s4 & s5 \\
\midrule
\multirow{12}{*}{\rotatebox[origin=c]{90}{WRN-28-10 / C100-C}} & \multirow{3}{*}{Deep Ens.} & Acc$\uparrow$ & 75.8 & 71.5 & 66.9 & 60.5 & 49.0 \\
 & & NLL$\downarrow$ & 0.942 & 1.129 & 1.349 & 1.722 & 2.501 \\
 & & ECE$\downarrow$ & 5.6 & 6.8 & 8.1 & 10.7 & 16.7 \\
\cmidrule(l){2-8}
 & \multirow{3}{*}{FiLM} & Acc$\uparrow$ & 75.5 & 71.0 & 66.2 & 59.6 & 48.6 \\
 & & NLL$\downarrow$ & 0.929 & 1.108 & 1.322 & 1.652 & 2.228 \\
 & & ECE$\downarrow$ & 4.4 & \textbf{3.9} & \textbf{4.2} & 5.6 & 8.6 \\
\cmidrule(l){2-8}
 & \multirow{3}{*}{Packed} & Acc$\uparrow$ & 66.9 & 61.8 & 56.8 & 50.2 & 39.3 \\
 & & NLL$\downarrow$ & 1.267 & 1.504 & 1.759 & 2.177 & 2.960 \\
 & & ECE$\downarrow$ & \textbf{4.3} & 5.8 & 7.6 & 10.8 & 17.2 \\
\cmidrule(l){2-8}
 & \multirow{3}{*}{$\sigma$N-Ens} & Acc$\uparrow$ & \textbf{78.0} & \textbf{74.1} & \textbf{69.8} & \textbf{63.2} & \textbf{52.6} \\
 & & NLL$\downarrow$ & \textbf{0.843} & \textbf{0.999} & \textbf{1.182} & \textbf{1.482} & \textbf{1.992} \\
 & & ECE$\downarrow$ & 5.4 & 5.0 & 4.8 & \textbf{5.5} & \textbf{6.5} \\
\midrule
\multirow{12}{*}{\rotatebox[origin=c]{90}{ResNet-50 / C100-C}} & \multirow{3}{*}{Deep Ens.} & Acc$\uparrow$ & \textbf{68.5} & \textbf{60.4} & \textbf{53.4} & \textbf{46.0} & \textbf{35.7} \\
 & & NLL$\downarrow$ & \textbf{1.318} & \textbf{1.719} & 2.112 & 2.595 & 3.280 \\
 & & ECE$\downarrow$ & 6.4 & 9.1 & 12.3 & 16.4 & 22.1 \\
\cmidrule(l){2-8}
 & \multirow{3}{*}{FiLM} & Acc$\uparrow$ & 66.6 & 58.3 & 51.4 & 43.8 & 33.9 \\
 & & NLL$\downarrow$ & 1.391 & 1.801 & 2.199 & 2.679 & 3.339 \\
 & & ECE$\downarrow$ & 7.2 & 10.2 & 13.5 & 17.8 & 23.7 \\
\cmidrule(l){2-8}
 & \multirow{3}{*}{Packed} & Acc$\uparrow$ & 67.7 & 59.5 & 52.6 & 45.4 & 35.4 \\
 & & NLL$\downarrow$ & \textbf{1.318} & 1.720 & \textbf{2.102} & \textbf{2.535} & \textbf{3.135} \\
 & & ECE$\downarrow$ & \textbf{4.6} & \textbf{7.1} & \textbf{10.0} & \textbf{13.4} & \textbf{18.3} \\
\cmidrule(l){2-8}
 & \multirow{3}{*}{$\sigma$N-Ens} & Acc$\uparrow$ & 67.5 & 59.0 & 52.0 & 44.5 & 34.5 \\
 & & NLL$\downarrow$ & 1.333 & 1.749 & 2.155 & 2.665 & 3.334 \\
 & & ECE$\downarrow$ & 4.8 & 7.5 & 10.6 & 14.8 & 20.2 \\
\bottomrule
\end{tabular}
\end{center}
\end{table}

\begin{table}[ht]
\caption{Distribution shift on the pretrained ViT-B/16 (ImageNet-C and CIFAR-100-C): per-severity accuracy, NLL, and ECE, each aggregated over the 15 corruption types ($M=4$, three seeds). Acc and ECE in \%; bold marks the best value per column within each setup.}
\label{tab:shift-vit}
\begin{center}
\small
\setlength{\tabcolsep}{5pt}
\begin{tabular}{ll l rrrrr}
\toprule
Setup & Method & Metric & s1 & s2 & s3 & s4 & s5 \\
\midrule
\multirow{12}{*}{\rotatebox[origin=c]{90}{ViT-B/16 / ImageNet-C}} & \multirow{3}{*}{Single} & Acc$\uparrow$ & 73.82 & 68.01 & 62.66 & 52.96 & 39.87 \\
 & & NLL$\downarrow$ & 1.043 & 1.367 & 1.666 & 2.257 & 3.190 \\
 & & ECE$\downarrow$ & 4.20 & 5.33 & 6.03 & 7.48 & 9.57 \\
\cmidrule(l){2-8}
 & \multirow{3}{*}{Deep Ens.} & Acc$\uparrow$ & 76.10 & 70.85 & 66.02 & 56.97 & 43.90 \\
 & & NLL$\downarrow$ & \textbf{0.923} & \textbf{1.201} & 1.459 & 1.975 & 2.833 \\
 & & ECE$\downarrow$ & \textbf{1.47} & 1.53 & 1.77 & 1.58 & \textbf{0.86} \\
\cmidrule(l){2-8}
 & \multirow{3}{*}{LoRA-Ens} & Acc$\uparrow$ & \textbf{76.33} & \textbf{71.37} & \textbf{66.94} & \textbf{58.80} & \textbf{46.03} \\
 & & NLL$\downarrow$ & 0.967 & 1.222 & \textbf{1.458} & \textbf{1.908} & \textbf{2.717} \\
 & & ECE$\downarrow$ & 6.73 & 6.86 & 7.04 & 6.78 & 4.70 \\
\cmidrule(l){2-8}
 & \multirow{3}{*}{$\sigma$N-Ens} & Acc$\uparrow$ & 74.85 & 69.23 & 63.95 & 54.53 & 41.45 \\
 & & NLL$\downarrow$ & 0.998 & 1.294 & 1.583 & 2.135 & 3.015 \\
 & & ECE$\downarrow$ & 1.72 & \textbf{1.46} & \textbf{1.35} & \textbf{0.96} & 1.69 \\
\midrule
\multirow{12}{*}{\rotatebox[origin=c]{90}{ViT-B/16 / CIFAR-100-C}} & \multirow{3}{*}{Single} & Acc$\uparrow$ & 86.58 & 82.07 & 77.71 & 71.31 & 61.34 \\
 & & NLL$\downarrow$ & 0.643 & 0.892 & 1.146 & 1.554 & 2.195 \\
 & & ECE$\downarrow$ & 8.17 & 10.96 & 13.60 & 17.48 & 23.63 \\
\cmidrule(l){2-8}
 & \multirow{3}{*}{Deep Ens.} & Acc$\uparrow$ & \textbf{87.39} & \textbf{83.12} & \textbf{78.94} & \textbf{72.69} & \textbf{62.82} \\
 & & NLL$\downarrow$ & 0.522 & 0.726 & 0.939 & 1.296 & 1.861 \\
 & & ECE$\downarrow$ & 5.37 & 7.18 & 9.03 & 12.07 & 17.06 \\
\cmidrule(l){2-8}
 & \multirow{3}{*}{LoRA-Ens} & Acc$\uparrow$ & 85.67 & 80.61 & 75.92 & 69.04 & 59.35 \\
 & & NLL$\downarrow$ & 0.580 & 0.779 & 0.983 & 1.302 & \textbf{1.757} \\
 & & ECE$\downarrow$ & 7.69 & 7.24 & 6.24 & \textbf{4.56} & \textbf{3.41} \\
\cmidrule(l){2-8}
 & \multirow{3}{*}{$\sigma$N-Ens} & Acc$\uparrow$ & 86.90 & 82.11 & 77.56 & 70.97 & 61.10 \\
 & & NLL$\downarrow$ & \textbf{0.503} & \textbf{0.707} & \textbf{0.922} & \textbf{1.266} & 1.800 \\
 & & ECE$\downarrow$ & \textbf{3.10} & \textbf{4.42} & \textbf{5.97} & 8.51 & 12.66 \\
\bottomrule
\end{tabular}
\end{center}
\end{table}

\end{document}